%% file: CTBNCToolkit.tex
\documentclass{article}

\usepackage[USenglish]{babel}
\usepackage{amsthm}
\usepackage{amsfonts}
\usepackage{amssymb}
\usepackage{amsmath}
\usepackage{graphicx}
\usepackage{rotating}
\usepackage{tabularx}
\usepackage{color}
\usepackage{microtype}
\usepackage{algorithmic}
\usepackage{algorithm}
\usepackage[round]{natbib} 
\usepackage{hyperref}
\usepackage{footmisc}
\usepackage{tikz}
\usetikzlibrary{shapes,arrows}	

\usepackage{fancyvrb}
\newcommand\code{\bgroup\@makeother\_\@makeother\~\@makeother\$\@codex}
\def\@codex#1{{\normalfont\ttfamily\hyphenchar\font=-1 #1}\egroup}
\let\code=\texttt
\let\proglang=\textsf
\DefineVerbatimEnvironment{Code}{Verbatim}{}
\DefineVerbatimEnvironment{CodeInput}{Verbatim}{fontshape=sl}
\DefineVerbatimEnvironment{CodeOutput}{Verbatim}{}
\newenvironment{CodeChunk}{}{}

\newcommand{\vc}[1]{\mathbf{#1}} 		
\newcommand{\s}[1]{\mathcal{#1}}		
\newcommand{\Pa}{Pa} 				
\newcommand{\pa}{pa} 				

\theoremstyle{plain}

\theoremstyle{definition}
\newtheorem{definition}{Definition}[section]
\newtheorem{example}{Example}[section]

\title{CTBNCToolkit: Continuous Time Bayesian Network Classifier Toolkit}
\author{Daniele Codecasa $\ddag$\\codecasa@disco.unimib.it\\DISCo,  Universit\`a degli Studi\\di Milano-Bicocca \and Fabio Stella $\ddag$\\stella@disco.unimib.it\\DISCo,  Universit\`a degli Studi\\di Milano-Bicocca}
\date{}

\begin{document}
\maketitle

\begin{abstract}
 Continuous time Bayesian network classifiers are designed for temporal classification of multivariate streaming data when time duration of events matters and the class does not change over time.
This paper introduces the CTBNCToolkit: an open source \proglang{Java} toolkit which provides a stand-alone application for temporal classification and a library for continuous time Bayesian network classifiers.
CTBNCToolkit implements the inference algorithm, the parameter learning algorithm, and the structural learning algorithm for continuous time Bayesian network classifiers. The structural learning algorithm is based on scoring functions: the marginal log-likelihood score and the conditional log-likelihood score are provided.
CTBNCToolkit provides also an implementation of the expectation maximization algorithm for clustering purpose.
The paper introduces continuous time Bayesian network classifiers.
How to use the CTBNToolkit from the command line is described in a specific section. Tutorial examples are included to facilitate users to understand how the toolkit must be used.
A section dedicate to the \proglang{Java} library is proposed to help further code extensions.
\end{abstract}

\subsection*{$\ddag$ Authors' contributions}
The toolkit was developed by Daniele Codecasa, who also wrote the paper. Fabio Stella read the
paper and made valuable suggestions. A former MATLAB implementation of
the CTBNC inference algorithm was made available by Fabio Stella in order to
test the correctness of the inference using the CTBNCToolkit\footnote{CTBNCToolkit website and the repository will be updated soon.}.
\newpage

\input{./intro.tex}

\input{./basicnotions.tex}

\input{./use.tex}

\input{./examples.tex}

\input{./library.tex}
\input{./conclusions.tex}



\bibliography{references}

\end{document}

%% file: intro.tex
\section{Introduction} \label{sec:Introduction}

The relevance of streaming data is increasing year after year with the emerging of new sources of data.
Data streams are important in engineering, with reference to image, audio and video processing \citep{yilmaz2006object} and in {\em computer science}, with reference to system error logs, web search query logs, network intrusion detection, and social networks \citep{Simma10}.
In finance, data streams are deeply studied for high frequency trading \citep{dacorogna2001introduction}. 
Biology and medicine offer other examples of analysis of data that change over time. Biologists analyze streaming data to model the evolution of infections \citep{barber2010graphical} and  to learn and analyze metabolic networks \citep{Voit2012}. While in medicine streaming data are used in patient monitoring based on sensor data and in continuous time diagnosis, including the study of computational firing pattern of neurons \citep{truccolo2005point}.

Data streaming problems may be approached with many algorithms and models. Among these, Dynamic Bayesian Networks (DBNs)\citep{dean1989model} and Hidden Markov Models (HMMs) \citep{rabiner1989tutorial} have received great attention for modeling temporal dependencies.
However, DBNs are concerned with discrete time and thus suffer from several limitations due to the fact that it is not clear how timestamps should be discretized. In the case where a too slow sampling rate is used the data will be poorly represented, while a too fast sampling rate rapidly makes learning and inference prohibitive. Furthermore, it has been pointed out \citep{NIPS2011_1106} that when allowing long term dependencies, it is necessary to condition on multiple steps into the past. Thus the choice of a too fast sampling rate will increase the number of such steps that need to be conditioned on.
Continuous Time Bayesian Networks (CTBNs) \citep{Nodelman+al:UAI05EP}, Continuous Time Noisy-OR (CT-NOR) \citep{SimmaGMBBIM08}, Poisson cascades \citep{Simma10} and Poisson networks \citep{rajaram2005poisson} together with the Piecewise-constant Conditional Intensity Model (PCIM) \citep{NIPS2011_1106} are interesting models to represent and analyze continuous time processes. {CT-NOR} and Poisson cascades are devoted to model event streams while they require the modeler to specify a parametric form for temporal dependencies. This aspect significantly impacts performance, and the problem of model selection in {CT-NOR} and Poisson cascades has not been addressed yet. This limitation is overcome by {PCIMs} which perform structure learning to model how events in the past affect future events of interest. {CTBNs} are continuous time homogeneous Markov processes which allow to represent joint trajectories of discrete finite variables, rather than models of event streams in continuous time.

Classification of data stream is an important area in data stream analysis with applications in many domains, such as medicine where the classification of multivariate trajectories is used for gesture recognition purpose related to the post-stroke rehabilitation \citep{ToGiQuSte2009}.
In this paper the problem of {\em temporal classification}, where data stream measurements are available over a period of time in history while the class is expected to occur in the future, is considered. 
Temporal classification can be addressed by discrete and continuous time models. Discrete time models include dynamic latent classification models \citep{zhong2012bayesian}, a specialization of the latent classification model (LCM) \citep{langseth2005latent}, and DBNs \citep{dean1989model}. Continuous time models, as Continuous Time Bayesian Network Classifiers (CTBNCs) \citep{stella2012continuous,codecasa13PKDD}, overcame the problem of timestamps discretization. 

This work describes the CTBNCToolkit: an open source  \proglang{Java} toolkit which allows temporal classification and clustering using the CTBNCs introduced by \citet{codecasa13PKDD,codecasa14}. CTBNCToolkit is a  \proglang{Java} library for CTBNCs which can be used as a stand-alone application through the command line interface.
CTBNCToolkit can be used for scientific purposes, such as model comparison and temporal classification of interesting scientific problems, but it can be used as well as a prototype to address real world problems.

CTBNCToolkit provides:
\begin{itemize}
\item the CTBNC inference algorithm \citep{stella2012continuous};
\item the CTBNC parameter learning algorithm \citep{stella2012continuous};
\item the scoring function based structural learning algorithm for CTBNCs \citep{codecasa13PKDD};
\item two scoring functions: the marginal log-likelihood score and the conditional log-likelihood score \citep{codecasa13PKDD};
\item the expectation maximization algorithm with soft and hard assignment for clustering purposes \citep{codecasa14};
\item three different validation methods: hold out, cross validation and a validation method for the clustering;
\item an extended set of performance measures for the supervised classification (see Section \ref{sec:results});
\item an extended set of external performance measures for clustering evaluation (see Section \ref{sec:results});
\item an extendable command line interface (see Section \ref{sec:experimentsrun}).
\end{itemize}

Section \ref{sec:basenotions} describes the basic notions about CTBNCs. A guide along the stand-alone usage of CTBNCToolkit is provided in Section \ref{sec:use}, which explains how to download the library (Section \ref{sec:download}), how to use the command line interface (Section \ref{sec:experimentsrun}), and how to read the classification and the clustering results (Section \ref{sec:results}).
Section \ref{sec:tutorial} provides tutorial examples of CTBNCtoolkit command line usage. All the examples are replicable.
The structure of the  \proglang{Java} library and the details of the implementation are addressed in Section \ref{sec:CTBNCLibrary}, while Section \ref{sec:conclusions} contains the conclusions and the possible future steps.

%% file: basicnotions.tex
\section{Basic notions}\label{sec:basenotions}
\subsection{Continuous time Bayesian networks}

Dynamic Bayesian networks (DBNs) model dynamical systems discretizing the time through several time slices. \citet{nodelman2002continuous} pointed out that ``{\em since DBNs slice time into fixed increments, one must always propagate the joint distribution over the variables at the same rate}" . Therefore, if the system consists of processes which evolve at different time granularities and/or the obtained observations are irregularly spaced in time,  the inference process may become computationally intractable.

Continuous time Bayesian networks (CTBNs) overcome the limitations of DBNs by explicitly representing temporal dynamics using exponential distributions.
Continuous time Bayesian networks (CTBNs) exploit the conditional independencies in continuous time Markov processes. A continuous time Bayesian network (CTBN) is a probabilistic graphical model whose nodes are associated with random variables and whose state evolves continuously over time.
\begin{definition} (Continuous time Bayesian network). \citep{nodelman2002continuous}.
Let $\vc{X}$ be a set of random variables $X_1, X_2, ..., X_N$. Each $X_n$ has a finite domain of values $Val(X_n)=\{x_1, x_2, ..., x_I \}$. A continuous time Bayesian network
$\aleph$ over $\vc{X}$ consists of two components:  the first is an initial distribution $P_{\vc{X}}^{0}$, specified as a Bayesian network $\s{B}$ over $\vc{X}$.
The second is a continuous transition model, specified as:
\begin{itemize}
	\item a directed (possibly cyclic) graph $\s{G}$ whose nodes are $X_1, X_2, ..., X_N$; $\Pa(X_n)$ denotes the parents of $X_n$ in $\s{G}$.
	\item a conditional intensity matrix, $\vc{Q}_{X_n}^{\Pa(X_n)}$, for each variable $X_n \in \vc{X}$.
\end{itemize}
\label{CTBN_definition}
\end{definition}
Given the random variable $X_n$, the {\em conditional intensity matrix} (CIM) $\vc{Q}_{X_n}^{\Pa(X_n)}$ consists of a set of intensity matrices, one intensity matrix
\[
\vc{Q}_{X_n}^{\pa(X_n)}=\left[ 
\begin{tabular}{ccccc}
$-q_{x_1}^{\pa(X_n)}$ 	     & $q_{x_1 x_2}^{\pa(X_n)} $ & . & $q_{x_1 x_I}^{\pa(X_n)}$ \\
$q_{x_2 x_1}^{\pa(X_n)}$ 	& $-q_{x_2}^{\pa(X_n)}$ 	 & . & $q_{x_2 x_I}^{\pa(X_n)}$ \\
		 . 		     & 			. 	      & . &  . \\
$q_{x_I x_1}^{\pa(X_n)}$ 	& $q_{x_I x_2}^{\pa(X_n)}$  & . &  $-q_{x_I}^{\pa(X_n)}$
\end{tabular}
\right] ,
\] 
\noindent for each instantiation $\pa(X_n)$ of the parents $\Pa(X_n)$ of node $X_n$, where $q_{x_i}^{\pa(X_n)}=\sum\limits_{x_j\neq x_i}q_{x_i x_j}^{\pa(X_n)}$ is the rate of leaving state $x_i$ for a specific instantiation $\pa(X_n)$ of $\Pa(X_n)$, while $q_{x_i x_j}^{\pa(X_n)}$ is the rate of arriving to state $x_j$ from state $x_i$ for a specific instantiation $\pa(X_n)$ of $\Pa(X_n)$.
Matrix $\vc{Q}_{X_n}^{\pa(X_n)}$ can equivalently be summarized by using two types of parameters, $q_{x_i}^{\pa(X_n)}$ which is associated with each state $x_i$ of the variable $X_n$ when its' parents are set to $\pa(X_n)$, and $\theta_{x_i x_j}^{\pa(X_n)}=\frac{q_{x_i x_j}^{\pa(X_n)}}{q_{x_i}^{\pa(X_n)}}$ which represents the probability of transitioning from state $x_i$ to state $x_j$, when it is known that the transition occurs at a given instant in time.

\begin{example}
Figure \ref{fig:eatingToy} shows a part of the drug network introduced in \citet{nodelman2002continuous}. It contains a cycle, indicating that whether a person is hungry ({\em H}) depends on how full his/her stomach ({\em S}) is, which depends on whether or not he/she is eating ({\em E}), which in turn depends on whether he/she is hungry.

Assume that $E$ and $H$ are binary variables with states {\em no} and {\em yes} while the variable $S$ can be in one of the following states; {\em full}, {\em average} or {\em empty}. Then, the variable $E$ is fully specified by the [2$\times$2] CIM matrices $\vc{Q}_E^{n}$, and $\vc{Q}_E^{y}$, the variable $S$ is fully specified by the [3$\times$3] CIM matrices $\vc{Q}_S^{n}$ and $\vc{Q}_S^{y}$, while the the variable $H$ is fully specified by the [2$\times$2] CIM matrices $\vc{Q}_H^{f}$, $\vc{Q}_H^{a}$ and, $\vc{Q}_H^{e}$.
\begin{figure}[h!]
	\centering
	\scalebox{0.7}{
	\begin{tikzpicture} [scale=.6,every text node part/.style={text centered},every node/.style={circle},text width=3em]
  		\node[draw,fill=white!20] (E) at (0,6) {Eating\\($E$)};
  		\node[draw,fill=white!20] (H) at (6,6) {Hungry\\($H$)};
  		\node[draw,fill=white!20] (S) at (3,1) {Full\\stomach\\($S$)};
		\foreach \from/\to in {H/E,E/S,S/H}
			\draw[->,thick] (\from) edge (\to);

  		\node (empty) at (18,1) {};

		\node (F1) at (9,7) {\begin{eqnarray}
			\vc{Q}_S^{y} & = & \left[
				\begin{array}{rrr}
					-q_{f}^{y} 			& \mbox{   }q_{f, a}^{y} 	& \mbox{   }q_{f, e}^{y} \\ 
					\mbox{   }q_{a, f}^{y} 	& -q_{a}^{y} 			& \mbox{   }q_{a, e}^{y} \\
					\mbox{   }q_{e, f}^{y} 	& \mbox{   }q_{e, a}^{y} 	& -q_{e}^{y} \\
				\end{array}
			      \right]\nonumber\\
			    & = &\left[
				\begin{array}{rrr}
					-0.03 & 0.02 & 0.01 \\ 
					5.99 & -6.00 & 0.01 \\
					1.00 & 5.00 & -6.00 \\
				\end{array}
			      \right]\hspace{2em}
			\label{eq:qToy}
			\end{eqnarray}};
		\node (F2) at (9,1) {\begin{eqnarray}
			\vc{Q}_S^y & = & \left[
				\begin{array}{rrr}
					q_{f}^{y} & 0 & 0 \\ 
					0 & q_{a}^{y} & 0 \\
					0 & 0 & q_{e}^{y}  \\
				\end{array}
			      \right]
			      \hspace{0.15cm} \left( 
				\left[
				\begin{array}{rrr}
					0 & \theta_{f, a}^{y} & \theta_{f, e}^{y} \\ 
					\theta_{a, f}^{y} & 0 & \theta_{a, e}^{y} \\
					\theta_{e, f}^{y} & \theta_{e, a}^{y} & 0 \\
				\end{array}
				 \right] 
				- \vc{I}
			      \right)\nonumber\\
			 & = &  \left[
				\begin{array}{rrr}
					0.03 & 0 & 0 \\ 
					0 & 6.00 & 0 \\
					0 & 0 & 6.00 \\
				\end{array}
			      \right]
			      \left(
				\left[
				\begin{array}{rrr}
					0                            & \frac{0.02}{0.03} & \frac{0.01}{0.03} \\ 
					\frac{5.99}{6.00} &  0                           & \frac{0.01}{6.00} \\
					\frac{1.00}{6.00} & \frac{5.00}{6.00} & 0 \\
				\end{array}
				 \right]- \vc{I}
			      \right)\hspace{2em}
		\label{eq:thetaToy}
		\end{eqnarray}};
	\end{tikzpicture}}
	\caption{A part of the drug network and the  two equivalent parametric representations of $\vc{Q}_S^{y}$ where $\vc{I}$ is the identity matrix.}
	\label{fig:eatingToy}
\end{figure}
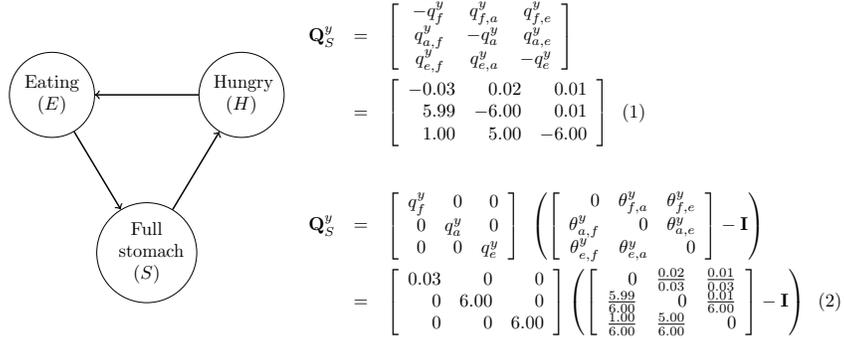

If the hours are the units of time, then a person who has an empty stomach ({\em S=empty}) and is eating ({\em E=yes}) is expected to stop having an empty stomach in $10$ minutes ($\frac{1.00}{6.00}$ hour). The stomach will then transition from state $empty$ ({\em S=empty}) to state $average$ ({\em S=average}) with probability $\frac{5.00}{6.00}$ and to state $full$ ({\em S=full}) with probability $\frac{1.00}{6.00}$.
Equation \ref{eq:qToy} is a compact representation of the CIM while Equation \ref{eq:thetaToy} is useful because it explicitly represents the transition probability value from state $x$ to state $x'$, i.e. $\theta_{xx^\prime}^{\pa(X)}$.
\end{example}

\subsection{Continuous time Bayesian network classifiers}
Continuous time Bayesian network classifiers (CTBNCs) \citep{stella2012continuous} are a specialization of CTBNs. CTBNCs allow polynomial time classification of a class variable that does not change over time, while general inference for CTBNs is NP-hard. Classifiers from this class explicitly represent the evolution in continuous time of the set of random variables $X_n$, $n=1, 2, ..., N$ which are assumed to depend on the static class node $Y$.
A continuous time Bayesian network classifier (CTBNC) is defined as follows.

\begin{definition} (Continuous time Bayesian network classifier). \citep{codecasa13PKDD}.
A continuous time Bayesian network classifier is a pair $\s{C}=\{\aleph, P(Y)\}$ where $\aleph$ is a CTBN model with attribute nodes $X_1, X_2, ..., X_N$, $Y$ is the class node with marginal
probability $P(Y)$ on states $Val(Y)=\{y_1, y_2, ..., y_K \}$, $\s{G}$ is the graph of the CTBNC, such that the following conditions hold: 
\begin{itemize}
	\item $\Pa(Y)=\emptyset$, the class variable $Y$ is associated with a root node;
	\item $Y$ is fully specified by $P(Y)$ and does not depend on time.
\end{itemize}
\label{CTBNC}
\end{definition}

\begin{example}
	Figure \ref{fig:eatingClassificationToy}a depicts the structure of a CTBNC to diagnose eating disorders from the eating process (Figure \ref{fig:eatingToy}). An example of the eating process is shown in Figure \ref{fig:eatingClassificationToy}b.
	\begin{figure}[h!]
		\centering
		\scalebox{0.65}{
		\begin{tikzpicture} [scale=.9,>=stealth',every text node part/.style={align=center},every node/.style={circle},text width=4em]

			\node (Ei1) at (5.1,11.1) {$\left\{\begin{array}{c} \mbox{ }\\ \{anorexia, bulimia, no\mbox{ }disorder\}\\ \mbox{ } \end{array}\right.$};
			\node[draw,fill=white!15] (D) at (3, 11) {Disorder\\($D$)};

			\node (Ei1) at (-0.1,7.6) {$\underbrace{\{yes,no\}}^{}$};
  			\node[draw,fill=white!15] (E) at (0,6) {Eating\\($E$)};

			\node (Hi1) at (6.1,7.6) {$\underbrace{\{yes,no\}}^{}$};
	  		\node[draw,fill=white!15] (H) at (6,6) {Hungry\\($H$)};

			\node (Si1) at (1.8,-0.8) {$\overbrace{\{full,average,empty\}}^{}$};
  			\node[draw,fill=white!15] (S) at (3,1) {Full\\stomach\\($S$)};

			\foreach \from/\to in {H/E,E/S,S/H}
				\draw[->,thick] (\from) edge (\to);
			\foreach \from/\to in {D/E,D/S,D/H}
				\draw[->,thick] (\from) edge (\to);
			\node[] (empty) at (-1,-2) {(a)};

			\node[] (empty) at (15-2,4.5) {(b)};
			\draw[->] (10.8-2.7,3+0) -- (16-2.7,3+0) node[right] {$t$};
			\draw[->] (11-2.7,3-0.2) -- (11-2.7,3+1) node[right] {$E$};
    			\draw[shift={(11-2.7,3+0)}] (2pt,0pt) -- (-2pt,0pt) node[left] {$n$};
    			\draw[shift={(11-2.7,3+1)}] (2pt,0pt) -- (-2pt,0pt) node[left] {$y$};
			\draw[thick] (11-2.7,3+1) -- (11.2-2.7,3+1);
			\draw[thick] (11.2-2.7,3+1) -- (11.2-2.7,3+0);
			\draw[thick] (11.2-2.7,3+0) -- (16-2.7,3+0);

			\draw[->] (10.8-2.7,0) -- (16-2.7,0) node[right] {$t$};
			\draw[->] (11-2.7,-0.2) -- (11-2.7,2) node[right] {$S$};
    			\draw[shift={(11-2.7,0)}] (1pt,0pt) -- (-2pt,0pt) node[left] {$e$};
    			\draw[shift={(11-2.7,1)}] (1pt,0pt) -- (-2pt,0pt) node[left] {$a$};
    			\draw[shift={(11-2.7,2)}] (1pt,0pt) -- (-2pt,0pt) node[left] {$f$};
			\draw[thick] (11-2.7,2) -- (11.6-2.7,2);
			\draw[thick] (11.6-2.7,2) -- (11.6-2.7,1);
			\draw[thick] (11.6-2.7,1) -- (13-2.7,1);
			\draw[thick] (13-2.7,1) -- (13-2.7,0);
			\draw[thick] (13-2.7,0) -- (16-2.7,0);

			\draw[->] (10.8-2.7,0-2) -- (16-2.7,0-2) node[right] {$t$};
			\draw[->] (11-2.7,-0.2-2) -- (11-2.7,1-2) node[right] {$H$};
    			\draw[shift={(11-2.7,0-2)}] (1pt,0pt) -- (-2pt,0pt) node[left] {$n$};
    			\draw[shift={(11-2.7,1-2)}] (1pt,0pt) -- (-2pt,0pt) node[left] {$y$};
			\draw[thick] (11-2.7,0-2) -- (14-2.7,0-2);
			\draw[thick] (14-2.7,0-2) -- (14-2.7,1-2);
			\draw[thick] (14-2.7,1-2) -- (16-2.7,1-2);
		\end{tikzpicture}}
	\caption{CTBNC to diagnose eating disorders (a) observing the eating process (b).}
	\label{fig:eatingClassificationToy}
	\end{figure}
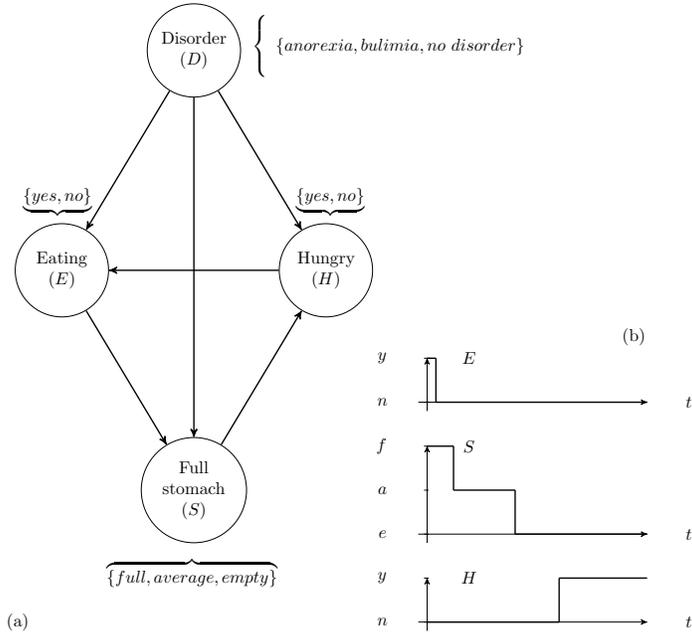
\end{example}

CTBNCs have been successfully used by \citet{codecasa13PKDD, codecasa14} in gesture recognition for post-stroke rehabilitation purposes \citep{ToGiQuSte2009}.

\subsubsection{Models}
\citet{stella2012continuous} defined the continuous time naive Bayes to overcome the effort of learning the structure of a CTBNC.
\begin{definition} (Continuous time naive Bayes classifier). \citep{stella2012continuous}.
A continuous time naive Bayes classifier is a continuous time Bayesian network classifier $\s{C}=\{\aleph, P(Y)\}$ such that $\Pa(X_n)=\{Y\}$, $n=1, 2, ..., N$.
\label{CTNB}
\end{definition}

While \citet{codecasa13PKDD} addressed for the first time the structural learning problem for CTBNCs defining two classifiers where the maximum number of parents (i.e. $k$) is bounded.
\begin{definition} (Max-$k$ Continuous Time Bayesian Network Classifier). \citep{codecasa13PKDD}.
A max-$k$ continuous time Bayesian network classifier is a couple $\s{M}=\{\s{C}, k\}$, where $\s{C}$ is a continuous time Bayesian network classifier $\s{C}=\{\aleph, P(Y)\}$ such that the number of parents $|\Pa(X_n)|$ for each attribute node $X_n$ is bounded by a positive integer $k$. Formally, the following condition holds; $|\Pa(X_n)| \leq k$, $n=1, 2, ..., N$, $k\geq 0$.
\label{KCTNBC}
\end{definition}

\begin{definition} (Max-$k$ Augmented Continuous Time Naive Bayes). \citep{codecasa13PKDD}.
A max-$k$ augmented continuous time naive Bayes classifier is a max-$k$ continuous time Bayesian network classifier
such that the class node $Y$ belongs to the parents set of each attribute node $X_n$, $n=1, 2, ..., N$. Formally, the following condition holds; $Y \in \Pa(X_n)$, $n=1, 2, ..., N$.
\label{ACTNB}
\end{definition}

\subsubsection{Parameter learning}\label{sec:paramslearning}
CTBNC parameter learning corresponds to CTBN parameter learning \citep{nodelman2002learning} with the exception of the class variable which has to be taken in account \citep{stella2012continuous,codecasa13PKDD}.

Given a data set $\s{D}$ and a fixed structure of a CTBNC, parameter learning is based on {\em marginal log-likelihood estimation}. Parameter learning accounts for the {\em imaginary counts}  of the hyperparameters $\alpha_x^{\pa(X)}$, $\alpha_{x x^\prime}^{\pa(X)}$ and, $\tau_x^{\pa(X)}$. 
The parameters $q_{x}^{\pa(X)}$ and $\theta_{x x^\prime}^{\pa(X_n)}$ can be estimated as follows:
\begin{itemize}
	\item $q_{x}^{\pa(X)} = \frac{\alpha_{x}^{\pa(X)} + M[x \mid \pa(X)]}{\tau_{x}^{\pa(X)} + T[x \mid \pa(X)]}$;
	\item $\theta_{x x^\prime}^{\pa(X)} = \frac{\alpha_{x x^\prime}^{\pa(X)} + M[x, x^\prime \mid \pa(X)]}{\alpha_{x}^{\pa(X)} + M[x \mid \pa(X)]}$.
\end{itemize}
where $M[x, x^\prime \mid \pa(X)]$, $M[x\mid \pa(X)]$ and  $T[x\mid \pa(X)]$ are the {\em sufficient statistics} computed over $\s{D}$:
\begin{itemize}
	\item $M[x, x^\prime \mid \pa(X)]$: number of times $X$ transitions from state $x$ to state $x^\prime$ when the state of its parents $\Pa(X)$ is set to $\pa(X)$;
	\item $M[x\mid \pa(X)] = \sum_{x^\prime\neq x} M[x, x^\prime \mid \pa(X)]$: number of transitions from state $x$ of variable $X$ when the state of its parents $\Pa(X)$ is set to $\pa(X)$;
	\item $T[x\mid \pa(X)]$: amount of time spent in state $x$ by variable $X$ when the state of its parents $\Pa(X)$ is set to $\pa(X)$.
\end{itemize}

The only difference between CTBNC parameter learning and CTBN parameter learning is related to the necessity of learning the probability distribution over the class node.
Because the class node is a static node, this can be done easily as follows:
\[
\theta_{y} = \frac{\alpha_y + M[y]}{\sum_{y^\prime}\alpha_{y^\prime} + M[y^\prime]}
\]
where $M[y]$ is the number of trajectories in the training set with class $Y$ sets to $y$ and, $\alpha_y$ are the imaginary counts related to the class variable.

\subsubsection{Structural learning}\label{sec:structlearning}
Learning a CTBNC from data consists of learning a CTBN where a specific node, i.e. the class node $Y$, does not depend on time.
In such a case, the learning algorithm runs, for each attribute node $X_n$, $n=1, 2, ..., N$, a local search procedure to find its optimal set of parents, i.e. the set of parents which maximizes a given score function. Furthermore, for each attribute node $X_n$, $n=1, 2, ..., N$, no more than $k$ parents are selected.

\citet{codecasa13PKDD} learned CTBNCs using the same local search algorithm proposed by \citet{nodelman2002learning} for CTBNs. \citet{codecasa13PKDD} proposed two possible scores: a marginal log-likelihood score, corresponding to the score introduced by \citet{nodelman2002learning}, and a conditional log-likelihood score, especially designed to improve the classification performances.
Because no closed form solution exists to compute the optimal value of the model's parameters when the conditional log-likelihood is used, \citet{codecasa13PKDD} followed the approach introduced and discussed by \citet{grossman2004learning} for static Bayesian classifiers. The scoring function is computed by using the conditional log-likelihood, while parameters values are obtained by using the Bayesian approach as described by \citet{nodelman2002learning}. 

The structural learning algorithm for CTBNCs and the conditional log-likelihood scoring function are described in \citep{codecasa13PKDD}.

\subsubsection{Classification}\label{sec:classification}
According to \citet{stella2012continuous} a CTBNC $\s{C}=\{\aleph, P(Y)\}$ classifies a stream of continuous time evidence $\vc{z}=(x_1, x_2, ..., x_N)$ for the attributes $\vc{Z}=(X_1, X_2, ..., X_N)$ over $J$ contiguous time intervals, i.e. a stream of continuous time evidence $\vc{Z}^{[t_1,t_2)}=\vc{z}^{[t_1,t_2)}$, $\vc{Z}^{[t_2,t_3)}=\vc{z}^{[t_2,t_3)}$, $\dots$, $\vc{Z}^{[t_{J-1},t_J)}=\vc{z}^{[t_{J-1},t_J)}$, by selecting the value $y^*$ for the class $Y$ which maximizes the posterior probability
\begin{equation}
P(Y| \vc{z}^{[t_1,t_2)}, \vc{z}^{[t_2,t_3)}, ..., \vc{z}^{[t_{J-1},t_J)} ),
\nonumber
\end{equation}
\noindent which is proportional to
\begin{equation}
P(Y) \prod_{j=1}^{J} q_{x^{j}_{m_j} x^{j+1}_{m_j}}^{\pa(X_{m_j})}  \prod_{n=1}^{N} exp \left(-q_{x^{j}_{n}}^{\pa(X_{n})} \delta_{j} \right),
\label{Aposteriori_rule}
\end{equation}
\noindent where:
\begin{itemize}
	\item $q_{x^{j}_{n}}^{\pa(X_{n})}$  is  the  parameter  associated  with  state $x^{j}_n$, in which the variable $X_n$ was  during  the $j^{th}$ time interval, given the state of  its parents $\pa(X_{n})$ during the $j^{th}$ time intervals;
	\item $q_{x^{j}_{m}x^{j+1}_{m}}^{\pa(X_{m})}$  is  the  parameter  associated  with  the  transition  from  state $x^{j}_m$, in which the variable $X_m$ was  during the $j^{th}$ time interval, to state $x^{j+1}_m$, in which  the  variable $X_m$ will be during the $(j+1)^{th}$ time interval, given the state of its parents $\pa(X_{m})$ during the $j^{th}$ and the $(j+1)^{th}$ time intervals,
\end{itemize}
\noindent while $\delta_{j}=t_{j}-t_{j-1}$ is the length of the $j^{th}$ time interval of the stream $\vc{z}^{[t_1,t_2)}, \vc{z}^{[t_2,t_3)}, ..., \vc{z}^{[t_{J-1},t_J)}$ of continuous time evidence.

The inference algorithm for CTBNCs (Algorithm \ref{algo:Classification_with_CTBNC}) is described in \citep{stella2012continuous}.
\begin{algorithm}[h!]
\caption{Inference algorithm for CTBNCs.}
\label{algo:Classification_with_CTBNC}
\begin{algorithmic}[1]
	\vspace{0.2cm}
		\REQUIRE a CTBNC $\mathcal{C}=\{\aleph, P(Y)\}$ consisting of $N$ attribute nodes and a class node $Y$ such that $Val(Y)=\{y_1, y_2, ..., y_K\}$, a {\em fully observed evidence stream}
			    $\left (\mathbf{x}^{1}, \mathbf{x}^{2}, ..., \mathbf{x}^{J} \right )$.
	\vspace{0.2cm}
\ENSURE  the maximum aposteriori classification $y^{*}$ for the {\em fully observed J-evidence-stream} $\left (\mathbf{x}^{1}, \mathbf{x}^{2}, ..., \mathbf{x}^{J} \right )$.
\\
	\vspace{0.2cm}
	\FOR{$k=1$ to $K$}
		\STATE $logp(y_k) \gets \log P(y_k)$
	\ENDFOR
	\vspace{0.2cm}
	\FOR{$k=1$ to $K$}

		\vspace{0.15cm}
		\FOR{$j=1$ to $J$}

			\vspace{0.1cm}
			\FOR{$n=1$ to $N$}

				\vspace{0.1cm}
					\STATE $logp(y_k) := logp(y_k) - q_{x^{j}_{n}}^{pa(X_n)} (t_j-t_{j-1})$

				\IF  {$x_{n}^{j} \neq x_{n}^{j+1}$ }
				\vspace{0.1cm}
					\STATE $logp(y_k) := logp(y_k) + log \left ( q_{x^{j}_{n}x^{j+1}_{n}}^{pa(X_n)} \right) $
				\vspace{0.1cm}
				\ENDIF 

			\vspace{0.1cm}
			\ENDFOR

		\vspace{0.1cm}
		\ENDFOR

	\vspace{0.1cm}
	\ENDFOR

	\vspace{0.2cm}
	\STATE $y^* \gets \arg\max_{y\in Val(Y)} logp(y)$.
	\vspace{0.2cm}

	\RETURN $y^*$

\end{algorithmic}
\end{algorithm}

\subsubsection{Clustering}\label{sec:CTBNCclustering}
Clustering for CTBNCs was introduced by \citet{codecasa14} using the {\em Expectation Maximization (EM)} algorithm.
The parameter learning algorithm relies on the same formulas showed in Section \ref{sec:paramslearning} for supervised learning, but where the expected sufficient statistics (i.e. $\bar{M}$ and $\bar{T}$) are used.

The expected sufficient statistics can be calculated in the EM expectation step by summing the contributions of the sufficient statistics, occurrences (i.e. ${M}^i$) and times (i.e. ${T}^i$) for each trajectory of the training set (i.e. $i\in\{1,\dots,\mid\s{D}\mid\}$) as follows.
 Contribution to the class count sufficient statistics for class $Y=y$ is:
	\[
		\bar{M}[y] = \bar{M}[y] + P(y\mid \vc{z}^{i,1},\dots,\vc{z}^{i,J})\mbox{.}
	\]
	Contribution to the occurrence count sufficient statistics of attribute $X_n$ such that $Y\in\Pa(X_n)$ and $Y=y$ is:
	\begin{align}
		\bar{M}[x_n, x_n^\prime \mid \pa( &X_n)] = \bar{M}[x_n, x_n^\prime \mid \pa(X_n)]\nonumber\\
										 & + M^i[x_n, x_n^\prime \mid \pa(X_n)/y]\cdot P(y\mid \vc{z}^{i,1},\dots,\vc{z}^{i,J})\mbox{,}\nonumber
	\end{align}
	where $M^i[x_n, x_n^\prime \mid \pa(X_n)/y]$ represents the number of times $X_n$ transitions from state $x_n$ to state $x_n^\prime$ in the $i^{th}$ trajectory when $X_n$'s parents without the class node (i.e. $\Pa(X_n)/Y$) are set to $\pa(X_n)/y$.
	Contribution to the time count sufficient statistics of attribute $X_n$ such that $Y\in\Pa(X_n)$ and $Y=y$ is:
	\begin{align}
		\bar{T}[x_n \mid \pa(X_n)] = 	& \bar{T}[x_n \mid \pa(X_n)]\nonumber\\
								& + T^i[x_n \mid \pa(X_n)/y]\cdot P(y\mid\vc{z}^{i,1},\dots,\vc{z}^{i,J})\mbox{,}\nonumber
	\end{align}
	where $T^i[x_n \mid \pa(X_n)/y]$ is the amount of time along the $i^{th}$ trajectory in which attribute $X_n$ is in state $x_n$ when its parents without the class node (i.e. $\Pa(X_n)/Y$) are set to $\pa(X_n)/y$.

The calculation of the expected sufficient statistics as showed before correspond to the soft assignment expectation step in the EM algorithm. CTBNCToolkit provides also the hard assignment EM algorithm \citep{kollar2009probabilistic} where the trajectory contributions to the expected sufficient statistics are calculated taking in account only the case of the most probable class.
Once calculated the expected sufficient statistics, the EM maximization step corresponds to the application of the formula in Section \ref{sec:paramslearning} using the expected sufficient statistics.

%% file: use.tex
\section{CTBNCToolkit how to}\label{sec:use}

\subsection{Download}\label{sec:download}
CTBNCToolkit can be downloaded as a stand-alone application. It requires {\em opencsv-2.3} library\footnote{\url{http://opencsv.sourceforge.net/}} to read the csv files and {\em commons-math3-3.0} library\footnote{\url{http://commons.apache.org/proper/commons-math/download_math.cgi}} for the Gamma function calculation.

To download and use the compiled CTBNCToolkit follow these steps:
\begin{itemize}
	\item download CTBNCToolkit \code{.jar} file from \url{http://dcodecasa.wordpress.com/ctbnc/ctbnctoolkit/} website;
	\item download opencsv library from \url{http://sourceforge.net/projects/opencsv/} web site (tests were made with version 2.3);
	\item download commons math library from \url{http://commons.apache.org/proper/commons-math/download_math.cgi} web site (tests were made with version 3.0).
\end{itemize}

The compiled CTBNCToolkit can be used as a stand-alone application, as showed in the next sections. On the same website where the \code{.jar} file is released, it is possible to find the published papers related to CTBNCs and a collection of free data sets to test CTBNCToolkit (see Section \ref{sec:tutorial}).

CTBNCToolkit source code is released under GPL v2.0\footnote{GPL v2.0 license: \url{http://www.gnu.org/licenses/gpl-2.0.html}} license. The source code is available on GitHub at the following url: \url{https://github.com/dcodecasa/CTBNCToolkit}.
It can be freely used in accordance with the GPL v2.0 license.

\subsection{Run experiments from the command line}\label{sec:experimentsrun}
Once downloaded the CTBNCToolkit jar file, or generated from the source code, it can be run as follows:
\begin{CodeChunk}
\begin{Code}
java -jar CTBNCToolkit.jar <parameters> <data>
\end{Code}
\end{CodeChunk}
where \code{<parameters>} are the CTBNCToolkit parameters (i.e modifiers), addressed in the following, and \code{<data>} is a directory containing a data set. This data set is used to generate both the training set and the test set (see Section \ref{sec:validation}), unless \code{-{}-training} or \code{-{}-testset} modifier are used. In this case \code{<data>} is considered the test set, and it can refer to a single file (see Section \ref{sec:training} and Section \ref{sec:testset}).
Hereafter the terms parameters and modifiers will be used interchangeably.

For sake of simplicity, the paths of the required libraries are defined in the manifest file contained in the \code{.jar} archive. Libraries are supposed to be saved in \code{lib/} directory. Tests are made using the \proglang{Java} virtual machine version 1.7.0\_25 in ubuntu.

It is worthwhile to note that it is often necessary to add the \proglang{Java} virtual machine parameter \code{-Xmx} to increase the heap space and to avoid out of memory exceptions (for example \code{-Xmx2048m} increases the heap dimension to 2Gb).

CTBNCToolkit parameters can be without any arguments, if specified as \code{-{}-modifier}; or with any number of arguments, if specified as \code{-{}-modifier=arg1,arg2,..,argN}.
Table \ref{tab:params} summarizes all the parameters while Table \ref{tab:params-comp} shows parameter incompatibilities and dependencies. The next sections address each parameter separately.
\begin{table}[h!]
	\centering
	\scalebox{0.9}{
	\addtolength{\tabcolsep}{-2pt}
	\begin{tabular}{|l|l|c|c|}
	\hline
		Parameter				&	Method	&	Arguments	&	Section\\
	\hline
	\hline
		\code{-{}-help} 			& \code{printHelp}			& no		& \ref{sec:help}\\
	\hline
		\code{-{}-CTBNC} 			& \code{setCTBNCModels}	& yes	& \ref{sec:CTBNC}\\
	\hline
		\code{-{}-model} 			& \code{setModels}			&  yes	& \ref{sec:model}\\
	\hline
		\code{-{}-validation} 		& \code{setValidationMethod}	&  yes	& \ref{sec:validation}\\
	\hline
		\code{-{}-clustering} 		& \code{setClustering}		& yes	& \ref{sec:clustering}\\
	\hline
		\code{-{}-1vs1} 			& \code{setModelToClass}	& no		& \ref{sec:1vs1}\\
	\hline
		\code{-{}-bThreshold} 		& \code{setBinaryDecider}	& yes	& \ref{sec:bThreshold}\\
	\hline
		\code{-{}-testName} 		& \code{setTestName}		& yes	& \ref{sec:testName}\\
	\hline
		\code{-{}-ext} 			& \code{setFileExt}			& yes	& \ref{sec:ext}\\
	\hline
		\code{-{}-sep} 			& \code{setFileSeparator}		& yes	& \ref{sec:sep}\\
	\hline
		\code{-{}-className} 		& \code{setClassColumnName}& yes	& \ref{sec:classClmName}\\
	\hline
		\code{-{}-timeName} 		& \code{setTimeColumnName}& yes	& \ref{sec:timeName}\\
	\hline
		\code{-{}-trjSeparator} 	& \code{setTrjSeparator}		& yes	& \ref{sec:trjSeparator}\\
	\hline
		\code{-{}-validColumns} 	& \code{setValidColumns}	& yes	& \ref{sec:validColumns}\\
	\hline
		\code{-{}-cvPartitions} 		& \code{setCVPartitions}		& yes	& \ref{sec:cvPartitions}\\
	\hline
		\code{-{}-cvPrefix} 		& \code{setCVPrefix}		& yes	& \ref{sec:cvPrefix}\\
	\hline
		\code{-{}-cutPercentage} 	& \code{setCutPercentage}	& yes	& \ref{sec:cutPercentage}\\
	\hline
		\code{-{}-timeFactor} 		& \code{setTimeFactor}		& yes	& \ref{sec:timeFactor}\\
	\hline
		\code{-{}-training} 		& \code{setTrainingSet}		&  yes	& \ref{sec:training}\\
	\hline
		\code{-{}-testset} 			& \code{setTestSet}			&  yes	& \ref{sec:testset}\\
	\hline
		\code{-{}-rPath} 			& \code{setResultsPath}		&  yes	& \ref{sec:rPath}\\
	\hline
		\code{-{}-confidence} 		& \code{setConfidence}		&  yes	& \ref{sec:confidence}\\
	\hline
		\code{-{}-noprob} 		& \code{disableProbabilities}	& no		& \ref{sec:noprob}\\
	\hline
		\code{-{}-v} 				& \code{setVerbose}			& no		& \ref{sec:v}\\
	\hline
	\end{tabular}}
	\caption{It is shown for each modifier: the method of the \code{CommandLine} class that manages it, the presence or absence of arguments, and the Section in which the modifier is described.}
	\label{tab:params}
\end{table}

\begin{table}[h!]
	\centering
	\scalebox{0.8}{
	\addtolength{\tabcolsep}{-1pt}
	\begin{tabular}{|@{}c@{}|@{}c@{}||p{0.25cm}|p{0.25cm}|p{0.25cm}|p{0.25cm}|p{0.25cm}|p{0.25cm}|p{0.25cm}|p{0.25cm}|p{0.25cm}|p{0.25cm}|p{0.25cm}|p{0.25cm}|p{0.25cm}|p{0.25cm}|p{0.25cm}|p{0.25cm}|p{0.25cm}|p{0.25cm}|p{0.25cm}|p{0.25cm}|p{0.25cm}|p{0.25cm}|p{0.25cm}|p{0.25cm}|}
	\hline
		Idx & Parameter 	& 1 & 2 & 3 & 4 & 5 & 6 & 7 & 8 & 9 & 10 & 11 & 12 & 13 & 14 & 15 & 16 & 17 & 18 & 19 & 20 & 21 & 22 & 23 & 24\\
	\hline
	\hline
		1 	& \code{-{}-help} 			& &X&X&X&X&X&X&X&X&X&X&X&X&X&X&X&X&X&X&X&X&X&X&X\\
	\hline
		2 	& \code{-{}-CTBNC} 		&X& & & & & & & & & & & & & & & & & & & & & & &\\
	\hline
		3 	& \code{-{}-model} 		&X& & & & &X& & & & &D& & &D& & & & & & & & & &\\
	\hline
		4 	& \code{-{}-validation} 		&X& & & &X& & & & & & & & & & & & & & & & & & &\\
	\hline
		5 	& \code{-{}-clustering} 		&X& & &X& &X&X& & & & & & & & & & & & & & & & &\\
	\hline
		6 	& \code{-{}-1vs1} 			&X& &X& &X& & & & & & & & & & & & & & & & & & &\\
	\hline
		7 	& \code{-{}-bThreshold} 	&X& & & &X& & & & & & & & & & & & & & & & & & &\\
	\hline
		8 	& \code{-{}-testName} 		&X& & & & & & & & & & & & & & & & & & & & & & &\\
	\hline
		9 	& \code{-{}-ext} 			&X& & & & & & & & & & & & & & & & & & & & & & &\\
	\hline
		10 	& \code{-{}-sep} 			&X& & & & & & & & & & & & & & & & & & & & & & &\\
	\hline
		11 	& \code{-{}-className} 	&X& & & & & & & & & & & & & & & & & & & & & & &\\
	\hline
		12 	& \code{-{}-timeName} 	&X& & & & & & & & & & & & & & & & & & & & & & &\\
	\hline
		13 	& \code{-{}-trjSeparator} 	&X& & & & & & & & & & & & & & & & & & & & & & &\\
	\hline
		14 	& \code{-{}-validColumns} 	&X& & & & & & & & & & & & & & & & & & & & & & &\\
	\hline
		15 	& \code{-{}-cvPartitions} 	&X& & &D& & & & & & & & & & & & & & & & & & & &\\
	\hline
		16 	& \code{-{}-cvPrefix} 		&X& & &D& & & & & & & & & & &D& & & & & & & & &\\
	\hline
		17 	& \code{-{}-cutPercentage} 	&X& & & & & & & & & & & & & & & & & & & & & & &\\
	\hline
		18 	& \code{-{}-timeFactor} 	&X& & & & & & & & & & & & & & & & & & & & & & &\\
	\hline
		19 	& \code{-{}-training} 		&X& & &D& & & & & & & & & & & & & & & & & & & &\\
	\hline
		20	& \code{-{}-testset}  		&X& &D&D& & & & & & & & & & & & & & &D& & & & &\\
	\hline
		21 	& \code{-{}-rPath} 		&X& & & & & & & & & & & & & & & & & & & & & & &\\
	\hline
		22 	& \code{-{}-confidence} 	&X& & & & & & & & & & & & & & & & & & & & & & &\\
	\hline
		23 	& \code{-{}-noprob} 		&X& & & & & & & & & & & & & & & & & & & & & & &\\
	\hline
		24 	& \code{-{}-v} 			&X& & & & & & & & & & & & & & & & & & & & & & &\\
	\hline
	\end{tabular}}
	\caption{Incompatibilities (X marks) and dependencies (D marks) between the modifiers. Incompatibility relations are symmetric; this does not hold for dependency relations, i.e. a modifier requires a particular value for another modifier, but the opposite it is not necessarily true. Modifiers are indicated by an index for due to the problems of space (first column).}
	\label{tab:params-comp}
\end{table}

\subsubsection{Help}\label{sec:help}
\code{-{}-help} prints on the screen the help that shows the allowed parameters. For each parameter a short description is provided.
When help is shown all the other parameters are ignored, and the program terminates after printing help.
\begin{CodeChunk}
\begin{Code}
java -jar CTBNCToolkit.jar --help
\end{Code}
\end{CodeChunk}

\subsubsection{Models to learn}\label{sec:CTBNC}
\code{-{}-CTBNC} modifier allows to specify the list of CTBNCs to test.

The models allowed follow:
\begin{itemize}
	\item \code{CTNB}: Continuous Time Naive Bayes (Definition \ref{CTNB}) \citep{stella2012continuous,codecasa13PKDD};
	\item \code{ACTNBk-f}: Max-$k$ Augmented Continuous Time Naive Bayes (Definition \ref{ACTNB}) \citep{codecasa13PKDD} where \code{k} is the number of parents ($\geq 2$) and \code{f} is the scoring function used to learn the structure (\code{LL} or \code{CLL});
	\item \code{CTBNCk-f}: Max-$k$ Continuous Time Bayesian Network Classifier (Definition \ref{KCTNBC}) \citep{codecasa13PKDD} where \code{k} is the number of parents ($\geq 1$) and \code{f} is the scoring function used to learn the structure (\code{LL} or \code{CLL});
\end{itemize}
where
\begin{itemize}
	\item \code{LL} stands for marginal log-likelihood scoring function \citep{nodelman2002learning,codecasa13PKDD};
	\item \code{CLL} stands for conditional log-likelihood scoring function \citep{codecasa13PKDD,codecasa14}.
\end{itemize}

After each model definition, it is possible to specify the modifier parameters, which define the imaginary counts of the hyperparameters (Section \ref{sec:paramslearning}):
\begin{itemize}
	\item \code{Mk}: \code{k} are the imaginary counts related to the number of transitions for each variable (default value: \code{1.0});
	\item \code{Tk}: \code{k} is the imaginary amount of time spent in a variable state (default value: \code{0.005});
	\item \code{Pk}: \code{k} are the imaginary counts related to the class occurrences (default value: \code{1.0}).
\end{itemize}
It is a good habit to avoid zero values for the imaginary counts. Indeed, when the data set is not big enough, some model parameters can be $0$ with the risk of a division by $0$ error. The parameters should be tuned in accord to the specific classification problem, since the parameters can have a big impact in the model performances.

In addition to each model learned with the structural learning, it is possible to add the following parameter:
\begin{itemize}
	\item \code{penalty}: adds the dimension penalty during the structural learning process; if omitted the penalty is disabled\footnote{The \code{penalty} flag is ignored if applied with the CTNB model.}.
\end{itemize}

Here is an example:
\begin{CodeChunk}
\begin{Code}
java -jar CTBNCToolkit.jar --CTBNC=CTNB,M0.1,T0.001,ACTNB2-CLL,CTBNC4-LL,
penalty <data>
\end{Code}
\end{CodeChunk}
The previous line enables the following tree models:
\begin{itemize}
	\item a CTNB with $0.1$ as prior for the variable counts, $0.001$ as the time prior and the standard $1.0$ as class imaginary counts;
	\item a Max-$2$ ACTNB learned maximizing the conditional log-likelihood score and with the default parameter priors;
	\item a Max-$4$ CTBNC learned maximizing the marginal log-likelihood, using the default priors and enabling the dimension penalty during the learning process.
\end{itemize}

\subsubsection{Model loading}\label{sec:model}
\code{-{}-model} modifier allows to specify the file paths of CTBNC models to load\footnote{The \code{-{}-model} modifier will be soon implemented.}.
The models must be in the \code{.ctbn} format (see Section \ref{sec:ctbnformat}).
If a training set is defined, the training set will be used only to learn models defined with the \code{-{}-CTBNC} modifier (see Section \ref{sec:CTBNC}).

Here are some examples:
\begin{itemize}
	\item tests the model stored in \code{model.ctbn} file:\\ \code{java -jar CTBNCToolkit.jar -{}-model=models/model.ctbn <other\_parameters>}\\ \code{<data>}
	\item tests the models stored in \code{model1.ctbn} and \code{model2.ctbn} files, and  learn and test a CTNB model:\\ \code{java -jar CTBNCToolkit.jar -{}-model=models/model1.ctbn,models/model2.ctbn}\\\code{-{}-CTBNC=CTNB <other\_parameters> <data>}
\end{itemize}

It is not possible to apply the \code{-{}-1vs1} modifier or to load models generated using the \code{-{}-1vs1} modifier (see Section \ref{sec:1vs1}).
The loaded models must be compatible with the input of \code{-{}-className} modifier (see Section \ref{sec:classClmName}) and \code{-{}-validColumns} modifier (see Section \ref{sec:validColumns}).

\subsubsection{ Validation method}\label{sec:validation}\index{classification!supervised}
The \code{-{}-validation} modifier allows to specify the validation method to use.
The CTBNCToolkit provides three validation methods: {\em hold-out}, {\em cross-validation} \citep{witten2005data} and a validation method used for the clustering tests.
The clustering validation method is automatically used when the clustering is enabled (see Section \ref{sec:clustering}).
While hold-out and cross-validation can be activated as follows:
\begin{itemize}
	\item \code{-{}-validation=HO,0.6}: enables the hold out validation method with a random partitioning of the data set in the training set ($60\%$) and test set ($40\%$) (default value: \code{0.7});
	\item \code{-{}-validation=CV,k}: enables the cross-validation with \code{k} folds (default value: \code{10}).
\end{itemize}

Here are some examples:
\begin{itemize}
	\item $70\%$-$30\%$ hold-out partitioning:
\begin{CodeChunk}
\begin{Code}
java -jar CTBNCToolkit.jar --validation=HO <other_parameters> <data>
\end{Code}
\end{CodeChunk}
	\item $60\%$-$40\%$ hold-out partitioning:
\begin{CodeChunk}
\begin{Code}
java -jar CTBNCToolkit.jar --validation=HO,0.6 <other_parameters> <data>
\end{Code}
\end{CodeChunk}
	\item \code{10}-folds cross-validation:
\begin{CodeChunk}
\begin{Code}
java -jar CTBNCToolkit.jar --validation=CV <other_parameters> <data>
\end{Code}
\end{CodeChunk}
	\item \code{8}-folds cross-validation:
\begin{CodeChunk}
\begin{Code}
java -jar CTBNCToolkit.jar --validation=CV,8 <other_parameters> <data>
\end{Code}
\end{CodeChunk}
\end{itemize}

No validation method can be specified in case of clustering (see Section \ref{sec:clustering}).

\subsubsection{Clustering}\label{sec:clustering}\index{classification!unsupervised}\index{clustering}\index{expectation maximization}\index{expectation maximization!soft-assignment}\index{expectation maximization!hard-assignment}
The \code{-{}-clustering} modifier disables the supervised learning and enables clustering \citep{codecasa14} (Section \ref{sec:CTBNCclustering}). 
Tests require labeled data sets because the performances are calculated over the complete data set using external measures (see Section \ref{sec:results_clust}), i.e. the {\em Rand index ($R$)} \citep{rand1971objective}, {\em Jaccard's coefficient ($J$)} \citep{halkidi2001clustering} and, the {\em Fowlkes-Mallows index ($FM$)} \citep{fowlkes1983method}.

CTBNC clustering is implemented using the Expectation Maximization (EM) algorithm \citep{kollar2009probabilistic}. Both {\em soft-assignment} and {\em hard-assignment} clustering are implemented (Section \ref{sec:CTBNCclustering}). The parameters allow to specify the clustering method, the termination criterion and the number of clusters as follows:
\begin{itemize}
	\item \code{hard}/\code{soft}: enables the clustering method  (default value: \code{soft});
	\item an integer number (i.e. \code{15}): sets to \code{15} the maximum number of iteration in the EM algorithm  (default value: \code{10});
	\item a \code{double} number (i.e. \code{0.1}): percentage of the data set trajectories; if less trajectories change class the EM algorithm is interrupted  (default value: \code{0.01});
	\item a \code{C} followed by an integer number (i.e. \code{C10}): sets the number of clusters (the default value is the number of classes of the class variable in the data set).
\end{itemize}
All these parameter values are optional and can be inserted in any order.

Here are some examples:
\begin{itemize}
	\item soft clustering, \code{10} iterations, $1\%$ as the trajectory threshold:
\begin{CodeChunk}
\begin{Code}
java -jar CTBNCToolkit.jar --clustering <other_parameters> <data>
\end{Code}
\end{CodeChunk}
	\item soft clustering, \code{15} iterations, $10\%$ as the trajectory threshold:
\begin{CodeChunk}
\begin{Code}
java -jar CTBNCToolkit.jar --clustering=soft,0.1,15 <other_parameters>
<data>
\end{Code}
\end{CodeChunk}
	\item soft clustering, \code{10} iterations, $3\%$ as the trajectory threshold:
\begin{CodeChunk}
\begin{Code}
java -jar CTBNCToolkit.jar clustering=0.03 <other_parameters> <data>
\end{Code}
\end{CodeChunk}
	\item hard clustering, \code{6} iterations, $5\%$ as the trajectory threshold, \code{5} clusters:
\begin{CodeChunk}
\begin{Code}
java -jar CTBNCToolkit.jar --clustering=6,0.05,hard,C5 <other_parameters>
<data>
\end{Code}
\end{CodeChunk}
\end{itemize}

When the clustering is enabled a dedicated validation method is used. The used validation method learns and tests the model on the whole data set.
 For this reason the validation modifier cannot be used (see Section \ref{sec:validation}). Also the classification threshold for binary class problem cannot be used (see Section \ref{sec:bThreshold}).

\subsubsection{One model one class}\label{sec:1vs1}
\code{-{}-1vs1} modifier enables the one model one class modality.
This modality generates for each model specified with the \code{-{}-CTBNC} modifier (see Section \ref{sec:CTBNC}) a set of models, one for each class. For example, if a CTNB is required using the \code{-{}-CTBNC} modifier over a $10$ class data set, this modifier will force the generation of $10$ CTNB models. Each one of the ten generated models will discriminate one class against the others.
During the classification process each model returns the probability related to the class for which it is specialized. The classification class is the one with the highest probability.

\begin{CodeChunk}
\begin{Code}
java -jar CTBNCToolkit.jar --1vs1 <other_parameters> <data>
\end{Code}
\end{CodeChunk}

\subsubsection{ Binary class threshold}\label{sec:bThreshold}
The \code{-{}-bThreshold} modifier changes the probability threshold used for assigning the class in supervised classification of binary problems.

Here are some examples:
\begin{itemize}
	\item the first class is assigned to a trajectory only if its posterior probability to belong to that class is greater or equal than \code{0.4}:
\begin{CodeChunk}
\begin{Code}
java -jar CTBNCToolkit.jar --bThreshold=0.4 <other_parameters> <data>
\end{Code}
\end{CodeChunk}
	\item the first class is assigned to a trajectory only if its posterior probability to belong to that class is greater or equal than \code{0.6}:
\begin{CodeChunk}
\begin{Code}
java -jar CTBNCToolkit.jar --bThreshold=0.6 <other_parameters> <data>
\end{Code}
\end{CodeChunk}
\end{itemize}
Classes are ordered alphabetically. For example, if ``A'' and ``B'' are the classes, the first example gives advantage to class ``A'', while the second example gives advantage to class ``B''.

This modifier can be used only for binary classification and cannot be used in the case of clustering (see Section \ref{sec:clustering}).

\subsubsection{Test name}\label{sec:testName}
The \code{-{}-testName} modifier specifies the name of the test. This name is used during the printing of the results to identify the particular test.
The default value is specified by the current time using the \code{"yyMMddHHmm\_Test"} format.

\begin{CodeChunk}
\begin{Code}
java -jar CTBNCToolkit.jar --testName=CTNBTest1 <other_parameters> <data>
\end{Code}
\end{CodeChunk}

\subsubsection{File extension}\label{sec:ext}
The \code{-{}-ext} modifier allows to specify the extension of the files to load in the data set directory (default value: \code{.csv}).

With the following command all the files in the data set directory with \code{.txt} extension are loaded:
\begin{CodeChunk}
\begin{Code}
java -jar CTBNCToolkit.jar --ext=.txt <other_parameters> <data>
\end{Code}
\end{CodeChunk}

\subsubsection{Column separator}\label{sec:sep}
The \code{-{}-sep} modifier allows to specify the column separator of the files to load (default value: \code{,}).

\begin{CodeChunk}
\begin{Code}
java -jar CTBNCToolkit.jar --sep=; <other_parameters> <data>
\end{Code}
\end{CodeChunk}

\subsubsection{Class column name}\label{sec:classClmName}
The \code{-{}-className} modifier specifies the name of the class column in the files to load (default value: \code{class}).

\begin{CodeChunk}
\begin{Code}
java -jar CTBNCToolkit.jar --className=weather <other_parameters> <data>
\end{Code}
\end{CodeChunk}

\subsubsection{Time column name}\label{sec:timeName}
The \code{-{}-timeName} modifier specifies the name of the time column in the files to load (default value: \code{t}).

\begin{CodeChunk}
\begin{Code}
java -jar CTBNCToolkit.jar --className=time <other_parameters> <data>
\end{Code}
\end{CodeChunk}

\subsubsection{Trajectory separator column name}\label{sec:trjSeparator}
Many data sets provide multiple trajectories in the same file. The \code{-{}-trjSeparator} enables trajectory separation using the information provided by a target column.
For example, if the trajectories are indexed by an incremental number (``trjIndex'' column) that indicates the trajectories in the file, it is possible to split the trajectories automatically, as follows:
\begin{CodeChunk}
\begin{Code}
java -jar CTBNCToolkit.jar --trjSeparator=trjIndex <other_parameters> <data>
\end{Code}
\end{CodeChunk}

By default this modifier is not used, and a one file - one trajectory matching is assumed.

\subsubsection{Data columns}\label{sec:validColumns}
The \code{-{}-validColumns} modifier allows to specify the columns of the files that correspond to the variables in the models to be generated. By default, i.e. when the modifier is not specified, all the columns except the time column (see Section \ref{sec:timeName}) and the trajectory separator column (see Section \ref{sec:trjSeparator}) are considered variables.

It is possible to specify any number of columns as arguments of the modifier. The following command line specifies the columns named \code{clmn1}, \code{clmn2}, and \code{clmn3} as the variables in the models to test:
\begin{CodeChunk}
\begin{Code}
java -jar CTBNCToolkit.jar --validColumns=clmn1,clmn2,clmn3 <other_parameters>
<data>
\end{Code}
\end{CodeChunk}

\subsubsection{Cross-validation partitions}\label{sec:cvPartitions}
The \code{-{}-cvPartitions} modifier allows to specify a partitioning when the cross-validation method is enabled (see Section \ref{sec:validation}).
This modifier indicates a file in which a cross-validation partitioning is specified. The cross-validation method will follow the partitioning instead of generating a new one.

The partitioning file can be a result file, generated by the CTBNCToolkit (see Section \ref{sec:results_class}):
\begin{CodeChunk}
\begin{Code}
Test1
trj12.txt: True Class: 4, Predicted: 4, Probability: 0.893885
...
trj87.txt: True Class: 2, Predicted: 2, Probability: 0.494983
Test2
trj26.txt: True Class: 2, Predicted: 2, Probability: 0.611254
...
trj96.txt: True Class: 2, Predicted: 3, Probability: 0.637652
Test3
trj1.txt: True Class: 4, Predicted: 4, Probability: 0.5697770
...
trj80.txt: True Class: 1, Predicted: 1, Probability: 0.938935
Test4
trj15.txt: True Class: 4, Predicted: 4, Probability: 0.624698
...
trj8.txt: True Class: 2, Predicted: 2, Probability: 0.7586410
Test5
trj11.txt: True Class: 4, Predicted: 4, Probability: 0.911368
...
trj99.txt: True Class: 4, Predicted: 4, Probability: 0.413442
\end{Code}
\end{CodeChunk}
or a text file where the word \code{Test} identifies the cross-validation folders, and the following lines specify the trajectories to load for each folder\footnote{The one trajectory - one file matching is supposed.}:
\begin{CodeChunk}
\begin{Code}
Test 1 of 5:
trj12.txt
...
trj87.txt
Test 2 of 5:
trj26.txt
...
trj96.txt
Test 3 of 5:
trj1.txt
...
trj80.txt
Test 4 of 5:
trj15.txt
...
trj8.txt
Test 5 of 5:
trj11.txt
...
trj99.txt
\end{Code}
\end{CodeChunk}

Here is an example where the file \code{partition.txt} in the \code{CV} directory is loaded:
\begin{CodeChunk}
\begin{Code}
java -jar CTBNCToolkit.jar --cvPartitions=CV/partition.txt <other_parameters>
<data>
\end{Code}
\end{CodeChunk}

This modifier requires to enable cross-validation (see Section \ref{sec:validation}).

\subsubsection{Cross-validation prefix}\label{sec:cvPrefix}
The \code{-{}-cvPrefix} allows to specify a prefix to remove from the trajectory names in the partition file.

For example, if the partition file loaded with the \code{-{}-cvPartitions} modifier has for some reasons the following form:
\begin{CodeChunk}
\begin{Code}
Test 1 of 5:
ex-trj12.txt
...
ex-trj87.txt
Test 2 of 5:
ex-trj26.txt
...
ex-trj96.txt
Test 3 of 5:
ex-trj1.txt
...
ex-trj80.txt
Test 4 of 5:
ex-trj15.txt
...
ex-trj8.txt
Test 5 of 5:
ex-trj11.txt
...
ex-trj99.txt
\end{Code}
\end{CodeChunk}
the \code{ex-} prefix can be removed automatically as follows:
\begin{CodeChunk}
\begin{Code}
java -jar CTBNCToolkit.jar --cvPrefix=ex- <other_parameters> <data>
\end{Code}
\end{CodeChunk}

This modifier requires the definition of a cross-validation partition file (see Section \ref{sec:cvPartitions}).

\subsubsection{Data sets reduction}\label{sec:cutPercentage}
The \code{-{}-cutPercentage} modifier allows to reduce the data set dimension in terms of number of trajectories and trajectory length.
With only one parameter (which is a percentage) it is possible to perform tests over reduced data sets in order to evaluate the ability of the models to work with a restricted amount of data.

The following line forces a random selection of $60\%$ of the data set trajectories and reduces each selected trajectory to $60\%$ of its original length:
\begin{CodeChunk}
\begin{Code}
java -jar CTBNCToolkit.jar --cutPercentage=0.6 <other_parameters> <data>
\end{Code}
\end{CodeChunk}

The default value is \code{1.0} that does not change the data amount.

\subsubsection{Time modifier}\label{sec:timeFactor}
The \code{-{}-timeFactor} modifier specifies a time factor used to scale the trajectory timing.

Here are some examples:
\begin{itemize}
	\item doubles the trajectory timing:
\begin{CodeChunk}
\begin{Code}
java -jar CTBNCToolkit.jar --timeFactor=2.0 <other_parameters> <data>
\end{Code}
\end{CodeChunk}
	\item reduces the trajectory timing by a factor of ten:
\begin{CodeChunk}
\begin{Code}
java -jar CTBNCToolkit.jar --timeFactor=0.1 <other_parameters> <data>
\end{Code}
\end{CodeChunk}
\end{itemize}

The default value is \code{1.0}, which implies no time transformations.

\subsubsection{Training set}\label{sec:training}
The \code{-{}-training} modifier specifies the directory that contains the training set.
Using this modifier the \code{<data>} path in the command line is used as the test set. If the \code{-{}-training} modifier is used, \code{<data>} can be a file or a directory.

Here are some examples:
\begin{itemize}
	\item the files in the \code{trainingDir/} directory compose the training set, while the files in \code{testDir/} compose the test set:
\begin{CodeChunk}
\begin{Code}
java -jar CTBNCToolkit.jar --training=trainingDir/ <other_parameters>
testDir/
\end{Code}
\end{CodeChunk}
	\item the files in the \code{training/} directory compose the training set, while the models are tested on the \code{testDir/trj.txt} file:
\begin{CodeChunk}
\begin{Code}
java -jar CTBNCToolkit.jar --training=training/ <other_parameters>
testDir/trj.txt
\end{Code}
\end{CodeChunk}
\end{itemize}

This modifier requires the use of the hold out validation method.

\subsubsection{Test set}\label{sec:testset}
The \code{-{}-testset} modifier forces to consider \code{<data>} as the test set\footnote{\code{-{}-testset} modifier will be soon implemented.}. \code{<data>} can be a folder or a file. 
This modifier can be used with the \code{-{}-model} modifier (see Section \ref{sec:model}) and the hold out validation method to avoid splitting the input data set into training and test set, when the definition of a training set is not necessary (i.e. when \code{-{}-CTBNC} modifier is not used). This modifier can be omitted, if \code{-{}-training} modifier (see Section \ref{sec:training}) is used.

Here are some examples:
\begin{itemize}
	\item the files in the \code{trainingDir/} directory compose the training set, while the files in \code{testDir/} compose the test set:
\begin{CodeChunk}
\begin{Code}
java -jar CTBNCToolkit.jar --training=trainingDir/ <other_parameters>
testDir/
\end{Code}
\end{CodeChunk}
	\item the files in the \code{trainingDir/} directory compose the training set, while the files in \code{testDir/} compose the test set:
\begin{CodeChunk}
\begin{Code}
java -jar CTBNCToolkit.jar --training=trainingDir/ --testset
<other_parameters> testDir/
\end{Code}
\end{CodeChunk}
	\item no training set is specified, \code{model.ctbn} is loaded and tested on \code{testDir/trj.txt} file:
\begin{CodeChunk}
\begin{Code}
\code{java -jar CTBNCToolkit.jar --testset --model=model.ctbn
<other\_parameters> testDir/trj.txt
\end{Code}
\end{CodeChunk}
\end{itemize}

This modifier requires the use of the hold out validation method.

\subsubsection{Results path}\label{sec:rPath}
The \code{-{}-rPath} modifier specifies the directory where to store the results (see Section \ref{sec:results}).
If it is not specified, the results are stored in a directory named as the test name (see Section \ref{sec:testName}) under the directory of the data (i.e. \code{<data>}) specified in the CTBNCToolkit command line.

Here are some examples:
\begin{itemize}
	\item save the results in \code{resultDir/} directory:
\begin{CodeChunk}
\begin{Code}
java -jar CTBNCToolkit.jar --rPath=resultDir --testName=Test1
<other_parameters> <data>
\end{Code}
\end{CodeChunk}
	\item save the results in \code{testDir/Test2/} directory:
\begin{CodeChunk}
\begin{Code}
java -jar CTBNCToolkit.jar --testName=Test2 <other_parameters> testDir/
\end{Code}
\end{CodeChunk}
	\item save the results in \code{testDir/Test3/} directory:
\begin{CodeChunk}
\begin{Code}
java -jar CTBNCToolkit.jar --testName=Test3 <other_parameters>
testDir/file.txt
\end{Code}
\end{CodeChunk}
\end{itemize}

\subsubsection{Confidence interval}\label{sec:confidence}
The \code{-{}-confidence} modifier allows to specify the confidence level to use in the model evaluation. The confidence level is used for model comparison in the case of supervised classification and for generating the error bars in the macro averaging curves (see Section \ref{sec:results_class}).
Models are compared using the test of difference of accuracy means proposed in \citep{witten2005data}. Supposes that two models $A$ and $B$ are compared with a test based on $90\%$ of confidence. If the test informs that the models are statistically different with $90\%$ of confidence, knowing that the mean accuracy of $A$ is greater than the mean accuracy of $B$ means that with $95\%$ of confidence model $A$ is better than model $B$ (for more details see  Section \ref{sec:results_class}). This should be taken into account when the confidence level is chosen.

The allowed levels of confidence are: $99.9\%$, $99.8\%$, $99\%$, $98\%$, $95\%$, $90\%$, and $80\%$. $90\%$ is the default value.

Here are some examples:
\begin{itemize}
	\item $90\%$ confidence:
\begin{CodeChunk}
\begin{Code}
java -jar CTBNCToolkit.jar <other_parameters> <data>
\end{Code}
\end{CodeChunk}
	\item $90\%$ confidence:
\begin{CodeChunk}
\begin{Code}
java -jar CTBNCToolkit.jar --confidence=90% <other_parameters> <data>
\end{Code}
\end{CodeChunk}
	\item $99\%$ confidence:
\begin{CodeChunk}
\begin{Code}
java -jar CTBNCToolkit.jar --confidence=99% <other_parameters> <data>
\end{Code}
\end{CodeChunk}
\end{itemize}

\subsubsection{Disable class probability}\label{sec:noprob}
The \code{-{}-noprob} modifier disables the calculation of the class probability distribution all across the trajectory.
The only probability values are calculated at the end of the trajectory. This allows to speed up the computation.

\begin{CodeChunk}
\begin{Code}
java -jar CTBNCToolkit.jar --noprob <other_parameters> <data>
\end{Code}
\end{CodeChunk}

\subsubsection{Verbose}\label{sec:v}
The \code{-{}-v} modifier enables the verbose modality. This modality prints more information that helps to follow the execution of the CTBNCToolkit.

\begin{CodeChunk}
\begin{Code}
java -jar CTBNCToolkit.jar --v <other_parameters> <data>
\end{Code}
\end{CodeChunk}

\subsection{Read the results}\label{sec:results}
Once a test is executed, the performances are printed in the results directory (see Section \ref{sec:rPath}).
The directory contains the performances of all the tested models.
The performances of each model are identified by its name. Model names start with \code{Mi}, where \code{i} is the index that identifies the model in the command line.

For example, the command line:
\begin{CodeChunk}
\begin{Code}
java -jar CTBNCToolkit.jar --CTBNC=CTNB,M0.1,T0.001,ACTNB2-CLL,CTBNC4-LL,
penalty <data>
\end{Code}
\end{CodeChunk}
generates the following model names: \code{M0\_CTNB}, \code{M1\_ACTNB2-CLL}, \code{M2\_CTBNC4-LL}.

In the case where external models are loaded (Section \ref{sec:model}), the \code{i} index first refers to the model defined with the \code{-{}-CTBNC} modifier (Section \ref{sec:CTBNC}), then it refers to the external models.

Here is another example. The command line:
\begin{CodeChunk}
\begin{Code}
java -jar CTBNCToolkit.jar --model=models/model1.ctbn,models/model2.ctbn
--CTBNC=CTNB <other_parameters> <data>
\end{Code}
\end{CodeChunk}
generates the following model names: \code{M0\_CTNB}, \code{M1\_model1}, \code{M2\_model2}.

The following sections describe the performances calculated in the case of classification and clustering. In both cases the results directory contains a file that shows the modifiers used in the test.

\subsubsection{Classification}\label{sec:results_class}

{\em Results file}

For each tested model a results file is provided. It can be found in the test directory and contains the results for each classified data set instance.
For each instance the name, the true class, the predicted class and the probability of the predicted class is shown.

Here is an example:
\begin{CodeChunk}
\begin{Code}
trj1: True Class: s2, Predicted: s2, Probability: 0.9942336124116266
trj10: True Class: s4, Predicted: s2, Probability: 0.9896614159579054
trj100: True Class: s1, Predicted: s1, Probability: 0.9955018513870293
trj11: True Class: s4, Predicted: s4, Probability: 0.977733953650775
trj12: True Class: s3, Predicted: s3, Probability: 0.9997240904249852
...
\end{Code}
\end{CodeChunk}

{\em Metrics}

In the results directory the metric file is stored. It is the \code{.csv} file that contains the following performances for each tested model \citep{witten2005data, fawcett2006introduction, japkowicz2011evaluating}:
\begin{itemize}
	\item Accuracy: accuracy value with the confidence interval calculated in accordance with the defined level of confidence (Section \ref{sec:confidence});
	\item Error: the percentage of wrong classified instances;
	\item Precision: precision value for each class;
	\item Recall: recall value for each class\footnote{Sensitivity, TP-rate and Recall are the same. \label{footnote:samemeasures}};
	\item F-Measure: balanced f-measure for each class (i.e. precision and recall weigh equally);
	\item PR AUC: precision-recall AUC (Area Under the Curve) for each class;
	\item Sensitivity: sensitivity for each class\footref{footnote:samemeasures};
	\item Specificity: specificity for each class;
	\item TP-Rate: true positive rate for each class\footref{footnote:samemeasures};
	\item FP-Rate: false positive rate for each class;
	\item ROC AUC: ROC Area Under the Curve for each class;
	\item Brier: brier value;
	\item Avg learning time: learning time or average learning time in case of multiple tests (i.e. cross-validation);
	\item Var learning time: learning time variance in case of multiple tests (i.e. cross-validation);
	\item Avg inference time: average inference time between all the tested instances;
	\item Var inference time: inference time variance between all the tested instances;
\end{itemize}
				
Some information in the metric file depends on the validation method. In the case of cross validation (Section \ref{sec:validation}), the file contains both the micro-averaging and the macro-averaging performances.\\

{\em Model comparison}

The model comparison file in the results directory contains the comparison matrices between the tested models.

A comparison matrix is a squared matrix that compares each pair of tested models by using their corresponding average accuracy values.
The comparison file contains the comparison matrices for the following confidence levels 99\%, 95\%, 90\%, 80\%, and 70\%.

Here is an example of the comparison matrix with 90\% confidence:\\

\scalebox{0.9}{
	\addtolength{\tabcolsep}{-2pt}
	\begin{tabular}{|c|c|c|c|c|c|c|c|}
	\hline
	Comparison test & 90\%&&&&&&\\
	\hline
	&M0&M1&M2&M3&M4&M5&M6\\
	\hline
	M0&&UP&UP&UP&UP&UP&$0$\\
	\hline
	M1&LF&&UP&$0$&UP&$0$&LF\\
	\hline
	M2&LF&LF&&LF&$0$&LF&LF\\
	\hline
	M3&LF&$0$&UP&&UP&$0$&LF\\
	\hline
	M4&LF&LF&$0$&LF&&LF&LF\\
	\hline
	M5&LF&$0$&UP&$0$&UP&&LF\\
	\hline
	M6&$0$&UP&UP&UP&UP&UP&\\
	\hline
\end{tabular}}\\

UP indicates that the upper model is statistically better than the left model, LF indicates that the left model is statistically better than the upper model, while $0$ indicates that the models are indistinguishable.

The tests are based on the difference of the average accuracy of the pair of considered models, as proposed in \citep{witten2005data}. This means that when UP or LF are showed we know that the two models are statistically different with a given confidence (i.e. $90\%$ in the previous table). Consider the previous table and the pair of models M0 and M1. We know that they are statistically different with $90\%$ of confidence. Because the average accuracy of M1 is bigger then the average accuracy of M0 the table shows that M1 is better than M0. Knowing that the test is a two tails test we actually know that M1 is statistically better of M0 with $95\%$ of confidence because only one tail must be considered.
This must be taken in account when the comparison matrices are read. \\

{\em Single run data}

The results directory contains a directory for each model. In the case of cross-validation, the model directory stores a directory named ``runs'' that contains a metric file with the performances of each single run.  In this directory all the models generated by the learning process  are stored.\\

{\em ROC curve}

The model directory also contains a directory named ``ROCs'' which stores the ROC curves for each class \citep{fawcett2006introduction}. In the case of cross-validation the curves calculated in micro and macro averaging are stored. The macro averaging curves show the confidence interval due to the vertical averaging. The confidence interval can be set using the \code{-{}-confidence} modifier (see Section \ref{sec:confidence}).\\

{\em Precision-Recall curve}

The model directory contains a directory named ``Precision-Recall'' which stores the precision-recall curves for each class  \citep{fawcett2006introduction}. In the case of cross-validation the curves calculated in micro and macro averaging are stored. The macro averaging curves show the confidence interval due to the vertical averaging. The confidence interval can be set using the \code{-{}-confidence} modifier (see Section \ref{sec:confidence}).\\

{\em Lift chart and cumulative response}

The directories of the models also contain a directory named ``CumulativeResp\&LiftChar'' which stores the cumulative response curves and the lift charts for each class \citep{fawcett2006introduction}. In the case of cross-validation the curves calculated in micro and macro averaging are stored. The macro averaging curves show the confidence interval due to the vertical averaging. The confidence interval can be set using the \code{-{}-confidence} modifier (see Section \ref{sec:confidence}).\\

\subsubsection{Clustering}\label{sec:results_clust}
In the case of clustering,  the results directory contains a directory for each tested model. Each directory contains a read me file with some information related to the test and the following files.\\

{\em Results file}

As in the case of supervised classification for each tested model a result file is provided. It contains the results for each classified data set instance.
For each instance the name, the true class, the predicted cluster and the probability of the predicted cluster is shown. Of course there is no correspondence between the class names and the cluster names.

Here is an example of the results file:
\begin{CodeChunk}
\begin{Code}
trj1.txt: True Class: 3, Predicted: 2, Probability: 0.9998733687935368
trj10.txt: True Class: 3, Predicted: 2, Probability: 0.9999799985527281
trj100.txt: True Class: 3, Predicted: 2, Probability: 0.9998951904099552
trj11.txt: True Class: 3, Predicted: 2, Probability: 0.9999999967591123
trj12.txt: True Class: 4, Predicted: 4, Probability: 0.9999999983049304
...
\end{Code}
\end{CodeChunk}

{\em Performances}

The file performances contains the following external measures \citep{halkidi2001clustering,gan2007data,xu2008clustering}:
\begin{itemize}
	\item the Rand index ($R$) \citep{rand1971objective};
	\item Jaccard's coefficient ($J$) \citep{halkidi2001clustering};
	\item the Fowlkes-Mallow index ($FM$) \citep{fowlkes1983method}.
\end{itemize}

In addition to these measures, association matrix, clustering-partition matrix, precision matrix, recall matrix, and f-measure matrix are shown. Learning time, average inference time and inference time variance are also shown.

When the number of clusters is set manually (see Section \ref{sec:clustering}) the performance file is not generated.\\

{\em Model file}

The model file shows the CTBNC learned in the test. Since the clustering validator learns one model all over the data set, just one model file is generated. \code{.ctbn} is the format used (see Section \ref{sec:ctbnformat}).

%% file: examples.tex
\section{Tutorial examples}\label{sec:tutorial}
This section provides replicable examples of CTBNCToolkit usage. 
To execute the tests follow the next steps:
\begin{itemize}
	\item download the compiled version of the CTBNCToolkit or download and compile the source code (see Section \ref{sec:download});
	\item create a directory for the tests where the compiled CTBNCToolkit must be copied (let's call the directory ``tutorial'');
	\item create a sub-directory named ``lib'' containing the required libraries (see Section \ref{sec:download})\footnote{Other directories can be used once the manifest file in the CTBNCToolkit\code{.jar}, is modified.};
	\item copy and unzip in the ``tutorial'' directory, for each of the following tests, the data sets used (see Section \ref{sec:download}).
\end{itemize}

\subsection{Classification}
In this section the tutorial examples of classification are given.
For these tests ``naiveBayes1'' synthetic data set is used. The data set has to be downloaded, unzipped and copied under the ``tutorial'' directory. The trajectories will be available in the ``tutorial/naiveBayes1/dataset/'' directory.

\subsubsection{Hold out}
Assume we are interested in making classifications using CTNB, ACTNB learned by maximizing conditional log-likelihood and CTBNC learned by maximizing marginal log-likelihood. The last two models learned setting to two the maximum number of parents. To evaluate the performances of the models we want to use the hold out validation method.

The data set is stored in \code{.txt} files. Trajectories use ``Class'' as the class column name and ``t'' as the time column name. To execute the test the following command must be used setting ``tutorial'' as the current directory\footnote{This test may takes a while.\label{footnote:takeawhile}}:
\begin{CodeChunk}
\begin{Code}
java -jar CTBNCToolkit.jar --CTBNC=CTNB,ACTNB2-CLL,CTBNC2-LL --validation=HO
--ext=.txt --className=Class naiveBayes1/dataset/
\end{Code}
\end{CodeChunk}
where:
\begin{itemize}
	\item \code{-{}-CTBNC=CTNB,ACTNB2-CLL,CTBNC2-LL} specifies the models to test (Section \ref{sec:CTBNC});
	\item \code{-{}-validation=HO} specifies hold out as the validation method (Section \ref{sec:validation});
	\item \code{-{}-ext=.txt} specifies that the data set files have \code{.txt} extension (Section \ref{sec:ext});
	\item \code{-{}-className=Class} specifies that the class column has the name ``Class'' (``class'' is the default value, see Section \ref{sec:classClmName});
	\item \code{naiveBayes1/dataset/} specifies the directory which contains the data set (Section \ref{sec:experimentsrun}).
\end{itemize}
It is worthwhile to notice that the time column name is not specified because ``t''  is the default name (Section \ref{sec:timeName}). Eventually the \code{-{}-v} modifier can be added to enable the verbose modality in order to monitor the program execution (Section \ref{sec:v}).

The results are stored in a directory with a name similar to ``131014\_1041\_Test''. To specify a different name use the \code{-{}-testName} modifier (Section \ref{sec:testName}).

The results follow the guidelines described in Section \ref{sec:results_class}.

\subsubsection{Cross validation loading a partitioning}
Assume we are interested in making classifications using just a CTNB. To evaluate the performances we use a $10$ fold cross-validation, loading a pre-defined partitioning. The partitions are described by the \code{NB-results.txt} file in the ``naiveBayes1/'' directory.

Similarly to the previous example, here is the command line to execute the test:
\begin{CodeChunk}
\begin{Code}
java -jar CTBNCToolkit.jar --CTBNC=CTNB --validation=CV
--cvPartitions=naiveBayes1/NB-results.txt --ext=.txt --className=Class
naiveBayes1/dataset/
\end{Code}
\end{CodeChunk}
The difference from the hold out classification example relies on the following two points:
\begin{itemize}
	\item \code{-{}-validation=CV} enables cross-validation to validate the models (Section \ref{sec:validation});
	\item \code{-{}-cvPartitions=naiveBayes1/NB-results.txt} specifies the file which contains the cross-validation partitioning (Section \ref{sec:cvPartitions}).
\end{itemize}
\code{NB-results.txt} is the results file (Section \ref{sec:results}) of a CTNB model learned on the same data set using the specified cross validation partitioning.
The results file can be loaded as a partitioning file (Section \ref{sec:cvPartitions}). This allows to replicate the executed tests. 

Because the test which generated the ``naiveBayes1/NB-results.txt'' result file was made using the same parameters, the same model and the same cross-validation partitioning, the results just obtained from the previous example should be exactly the same. This can be seen comparing the ``naiveBayes1/NB-results.txt'' file with the results file just generated.

\subsection{Clustering}
In this section a tutorial example of clustering is described.
Also in this case the ``naiveBayes1'' synthetic data set is used. The data set has to be downloaded, unzipped and copied under the ``tutorial'' directory. The trajectories will be contained in ``tutorial/naiveBayes1/dataset/'' directory.

Assume we are interested in making clustering using CTNB and CTBNC learned by maximizing marginal log-likelihood when the maximum number of parents is set to two.

The data set is stored in \code{.txt} files. Trajectories use ``Class'' as the class column name and ``t'' as the time column name. To execute the test the following command must be used setting ``tutorial'' as the current directory\footref{footnote:takeawhile}:
\begin{CodeChunk}
\begin{Code}
java -jar CTBNCToolkit.jar --CTBNC=CTNB,CTBNC2-LL --clustering --ext=.txt
--className=Class naiveBayes1/dataset/
\end{Code}
\end{CodeChunk}
The command is similar to the previous ones, the main difference is the use of the \code{-{}-clustering} modifier which substitutes the validation modifier (Section \ref{sec:clustering}).

The results are stored in a directory with a name similar to ``131014\_1252\_Test''. To specify a different name use the \code{-{}-testName} modifier (Section \ref{sec:testName}).

The results follow the guidelines described in Section \ref{sec:results_clust}.
Since learning and testing in case of clustering are done using the whole data set, it is easy to replicate the clustering experiments, but because of the random instantiation of the EM algorithm the results can be quite different especially if there is not a clear distinction between clusters.

%% file: library.tex
\section{CTBNCToolkit library} \label{sec:CTBNCLibrary}

CTBNCToolkit is released under the GPL v2.0\footnote{GPL v2.0 license: \url{http://www.gnu.org/licenses/gpl-2.0.html}} license.
It is available on GitHub at the following url: \url{https://github.com/dcodecasa/CTBNCToolkit}. See section \ref{sec:download} for more information regarding the CTBNCToolkit download.

The CTBNCToolkit is a library to manage CTBNCs. It provides:
\begin{itemize}
	\item a CTBNC model representation which can be easily extended to define other types of models (Section \ref{sec:models-code});
	\item a supervised learning algorithm (Section \ref{sec:learning-code});
	\item soft and hard assignment Expectation Maximization (EM) algorithms for clustering purposes (Section \ref{sec:clustering-code});
	\item two different scoring functions to learn CTBNCs (Section \ref{sec:learning-code});
	\item CTBNCs inference algorithm for static classification (Section \ref{sec:inference-code});
	\item three different validation methods to realize experiments (Section \ref{sec:validation-code}),
	\item a rich set of performance measures for supervised classification and clustering as well (Section \ref{sec:performances-code});
	\item a set of utilities for data set and experiment managing (Sections \ref{sec:input-code} and \ref{sec:tests-code});
	\item an extendable command line front-end (Section \ref{sec:frontend-code}).
\end{itemize}

In the following part of this section the main software components are analyzed from the development point of view.

\subsection{Input trajectories}\label{sec:input-code}

\begin{figure}[h!]	
	\centering
		\includegraphics[width=.85 \textwidth]{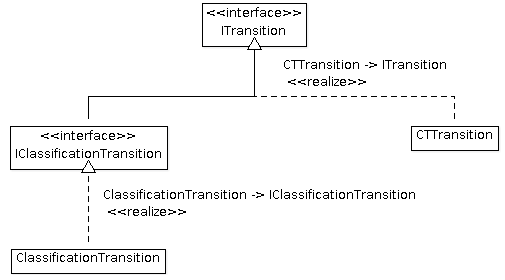}{\tiny(a)}\\
		\includegraphics[width=.85 \textwidth]{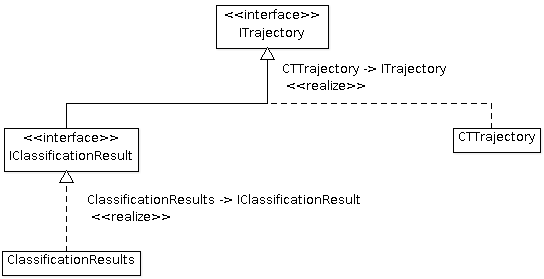}{\tiny(b)}
	\caption{Simplified class diagram of the trajectory managing components.}
	\label{fig:trajectories-class}
\end{figure}

Figure \ref{fig:trajectories-class} depicts the simplified class diagram of interfaces and classes used for representing trajectories.

\code{ITrajectory} is the interface that defines the trajectories. Each trajectory consists of a sequence of transitions. Each transition (i.e. \code{ITransition}) occurs at a particular time. Transition time is generalized using \code{<TimeType extends Number>} generic. Since CTBNCs model continuous time trajectories, in the stand-alone CTBNCToolkit implementation the \code{TimeType} generic is filled with \code{Double} class. Nevertheless, each extension of \code{Number} and \code{Comparable} classes can be potentially used as time representation; i.e. the \code{Integer} class can be specified to represent discrete time trajectories.

\code{CTTransition} and \code{CTTrajectory} provide a standard implementation of \code{ITransition} and \code{ITrajectory}.

\code{IClassificationTransition} and \code{IClassificationResult} interfaces define an extension of \code{ITransition} and \code{ITrajectory} which allows to mange classified trajectories.  They allow to insert the probability distribution for each transition into a trajectory.

\code{ClassificationTransition} and \code{ClassificationResults} implement a standard version of the interfaces for the classified transitions and trajectories.

\begin{example}
Here is how to generate a trajectory.
\begin{CodeChunk}
\begin{Code}
0.0	Va12	Vb1	Vc123
0.2	Va12	Vb2	Vc123
0.6	Va3	 Vb3	Vc123
1.3	Va4	 Vb4	Vc4
\end{Code}
\end{CodeChunk}
The previous trajectory shows multiple changes of states for each time instant, even if theoretically, in accord with the CTBN definition, this cannot happen.
This can be implemented by the following code:
\begin{CodeChunk}
\begin{Code}
// Nodes-column name definition
String[] nodeNames = new String[3];
nodeNames[0] = "A"; nodeNames[1] = "B"; nodeNames[2] = "C";
NodeIndexing nodeIndexing = NodeIndexing.getNodeIndexing(
	"IndexingName", nodeNames, nodeNames[0], null);
	
// Time jump definition
String[] v;
List<Double> times = new Vector<Double>();
List<String[]> values = new Vector<String[]>();
times.add(0.0);times.add(0.2);times.add(0.6);times.add(1.3);

// State definition
v = new String[3];
v[0] = "Va12"; v[1] = "Vb1"; v[2] = "Vc123";
values.add(v);
v = new String[3];
v[0] = "Va12"; v[1] = "Vb2"; v[2] = "Vc123";
values.add(v);
v = new String[3];
v[0] = "Va3"; v[1] = "Vb3"; v[2] = "Vc123";
values.add(v);
v = new String[3];
v[0] = "Va4"; v[1] = "Vb4"; v[2] = "Vc4";
values.add(v);
	
// Trajectory creation
CTTrajectory<Double> tr = new CTTrajectory<Double>(nodeIndexing, times,
	values);
\end{Code}
\end{CodeChunk}
Section \ref{sec:nodeIndexing} provides a clarification about \code{NodeIndexing} class.
\end{example}

The \code{CTBNCTestFactory} class, a class used in test management (see Section \ref{sec:tests-code}), provides a set of \code{static} methods which are useful for dealing with data sets:
\begin{itemize}
	\item \code{loadDataset}: loads a data set;
	\item \code{partitionDataset}: partitions a data set in accordance with a partitioning file (i.e. cross-validation folding, Section \ref{sec:cvPartitions});
	\item \code{loadResultsDataset}: loads a data set composed of classified trajectories; for each trajectory the probability distribution of the class has to be specified;
	\item \code{partitionResultDataset}: partitions a data set of results in accordance with a partitioning file (i.e. cross-validation folding);
	\item \code{permuteDataset}: randomly permutes a data set;
	\item \code{cutDataset}: reduces the original data set in terms of number and length of trajectories (Section \ref{sec:cutPercentage}).
\end{itemize}

\subsection{Global node indexing}\label{sec:nodeIndexing}
Before introducing the models in section \ref{sec:models-code}, it is necessary to deal with the global node indexing system.
Models nodes, variables, and columns in a trajectory use the same names.
Indeed, objects like model nodes or state values in a trajectory have to be stored by using the same names. To allow a really efficient method to recover these objects the \code{NodeIndexing} class is developed.
A model node can be recovered using its name, which corresponds to a column name in the trajectories, or using its index. Recovering objects by name uses a tree structure. This is efficient, but it is possible to improve it. The index solution is developed to be an $O(1)$ entity recovery system.

To guarantee the same indexing for all the trajectory columns and all the model nodes, the class \code{NodeIndexing} is used.
The idea is that each trajectory and each model has to be synchronized with the same \code{NodeIndexing} instance. To ensure this, \code{NodeIndexing} provides a static method (i.e. \code{getNodeIndexing}) which allows to obtain or create a \code{NodeIndexing} instance using an unique name. This name usually corresponds to the test name.

\begin{example}
Here is an example of how to create a new \code{NodeIndexing} class.
\begin{CodeChunk}
\begin{Code}
String[] nodeNames = new String[3]; nodeNames[0] = "class";
nodeNames[1] = "N1"; nodeNames[2] = "N2";
NodeIndexing nodeIndexing = NodeIndexing.getNodeIndexing("IndexingName",
	nodeNames, nodeNames[0], null);
\end{Code}
\end{CodeChunk}
The first argument is the unique key associated with the index. Using the same key in successive calls of the \code{getNodeIndexing} method allows to recover the same indexing instance in accordance with the singleton pattern.
The second argument (i.e. \code{nodeNames}) is the array of all the node names, while the third argument is the class node name.
The last argument is the set of all the node names in the \code{nodeNames} array to which an index will be associated. This argument allows to select just a subset of the names specified in the second argument. If it is left to \code{null}, all the names are loaded.
\end{example}

When two objects are synchronized with the same \code{NodeIndexing} instance, they use the same loaded indexing. This allows the direct communication index-to-index between all the objects with the same synchronization.

\subsection{CTBNCToolkit models}\label{sec:models-code}

\begin{figure}[h!]	
	\centering
		\includegraphics[width=0.85 \textwidth]{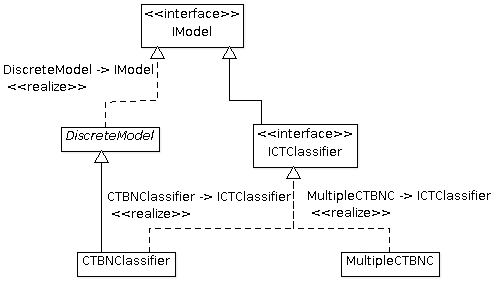}{\tiny(a)}\\
		\includegraphics[width=0.64 \textwidth]{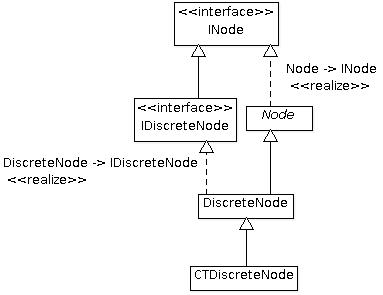}{\tiny(b)}
	\caption{Simplified class diagram of the model managing components.}
	\label{fig:models-class}
\end{figure}

Figure \ref{fig:models-class} depicts the simplified class diagram of the interfaces and classes to be used to work with the models.

\subsubsection{Nodes}
A model consists of nodes which represent the model variables. Node objects are shown in Figure \ref{fig:models-class}b.
\code{INode} interface defines the generic properties of a node, while \code{IDiscreteNode} interface defines the properties of a discrete state space node.

\code{Node} class is an abstract class which implements the properties that all the nodes require. Those properties are mainly related to the naming and the dependency relations between nodes.
\code{DiscreteNode} class implements all the requirements of a discrete node. It does not specify any quantitative component as Conditional Intensity Matrix (CIM) or Conditional Probability Table (CPT), but only manages the discrete states of a variable.

\code{CTDiscreteNode} class implements all the properties which the CTBNC nodes require. It allows the CIMs definition and can implement both a continuous time node and a static node for continuous time models (i.e. the class node).

\begin{example}
Here is an example of how to create a continuous time discrete node (i.e. \code{CTDiscreteNode}).
\begin{CodeChunk}
\begin{Code}
// State definition
Set<String> states2;Set<String> states3;
states2 = new TreeSet<String>();
states2.add("n1_1");states2.add("n1_2");
states3 = new TreeSet<String>();
states3.add("n1_1");states3.add("n1_2");states3.add("n1_3");

// Node creation
CTDiscreteNode node = new CTDiscreteNode("node",states3,false);
CTDiscreteNode parent = new CTDiscreteNode("parent",states2,true);
parent.addChild(node);
		
// CIM definition
double[][] cim = new double[3][3];
cim[0][0] = -2; cim[0][1] = 1; cim[0][2] = 1;
cim[1][0] = 2; cim[1][1] = -4; cim[1][2] = 2;
cim[2][0] = 2; cim[2][1] = 1; cim[2][2] = -3;
node.setCIM(0, cim);
node.setCIM(1, cim);
assertTrue( node.checkCIMs() == -1);
\end{Code}
\end{CodeChunk}
\end{example}

\subsubsection{Models}
Generic model properties are defined by the \code{IModel} interface. Part of these properties are implemented by the \code{DiscreteModel} abstract class.
Instead, the \code{ICTClassifier} interface defines the properties related to the classification process, which a continuous time classifier has to satisfy.

CTBNCs are implemented by \code{CTBNClassifier} class. While the \code{MultipleCTBNC} class implements a continuous time classifier composed of a set of binary CTBNCs, each one is specialized to recognize a class against the others (see Section \ref{sec:1vs1}).

Models are simple \proglang{Java} classes used like containers of nodes. The real complexity related to the learning and the classification processes relies on the algorithm classes.

\begin{example}
Here is an example that shows how to create a \code{CTBNClassifier}.
\begin{CodeChunk}
\begin{Code}
// Node names and indexing definition
String[] nodesNames = new String[4];
nodesNames[0] = "Class"; nodesNames[1] = "A";
nodesNames[2] = "B"; nodesNames[3] = "C";
NodeIndexing nodeIndexing = NodeIndexing.getNodeIndexing("IndexingName",
	nodesNames, nodesNames[0], null);
		
// State generation
CTDiscreteNode classNode, aNode, bNode, cNode;
Set<String> states2 = new TreeSet<String>();
Set<String> states3 = new TreeSet<String>();
Set<CTDiscreteNode> nodes = new TreeSet<CTDiscreteNode>();
states2.add("s1");states2.add("s2");
states3.add("s1");states3.add("s2");states3.add("s3");

// Node generation
nodes.add(classNode = new CTDiscreteNode(nodesNames[0], states2, true));
nodes.add(aNode = new CTDiscreteNode(nodesNames[1], states2, false));
nodes.add(bNode = new CTDiscreteNode(nodesNames[2], states3, false));
nodes.add(cNode = new CTDiscreteNode(nodesNames[3], states3, false));

// Model structure definition
classNode.addChild(aNode);
classNode.addChild(bNode);
classNode.addChild(cNode);

// Node CIM definition
double[][] cim;
cim = new double[1][2]; cim[0][0] = 0.5; cim[0][1] = 0.5;
classNode.setCIM(0, cim);
assertTrue(classNode.checkCIMs() == -1);

cim = new double[2][2]; 
cim[0][0] = -0.1; cim[0][1] = 0.1;
cim[1][0] = 0.1; cim[1][1] = -0.1; 
aNode.setCIM(0, cim);
cim = new double[2][2]; 
cim[0][0] = -5; cim[0][1] = 5;
cim[1][0] = 5; cim[1][1] = -5;
aNode.setCIM(1, cim);
assertTrue(aNode.checkCIMs() == -1);

cim = new double[3][3]; 
cim[0][0] = -0.7; cim[0][1] = 0.5; cim[0][2] = 0.2; 
cim[1][0] = 1.0; cim[1][1] = -1.6; cim[1][2] = 0.6; 
cim[2][0] = 2; cim[2][1] = 1.3; cim[2][2] = -3.3;
bNode.setCIM(0, cim);cNode.setCIM(0, cim);
cim = new double[3][3]; 
cim[2][2] = -0.7; cim[2][1] = 0.5; cim[2][0] = 0.2; 
cim[1][0] = 1.0; cim[1][1] = -1.6; cim[1][2] = 0.6; 
cim[0][2] = 2; cim[0][1] = 1.3; cim[0][0] = -3.3;
bNode.setCIM(1, cim); cNode.setCIM(1, cim);
assertTrue(bNode.checkCIMs() == -1);
assertTrue(cNode.checkCIMs() == -1);

// Model generation
CTBNClassifier model = new CTBNClassifier(nodeIndexing, "classifier", nodes);
\end{Code}
\end{CodeChunk}
\end{example}

\subsubsection{.ctbn format}\label{sec:ctbnformat}
\code{.ctbn} is the space-separated text format in which the CTBNCToolkit saves the learned models. The \code{toString()} method in the \code{CTBNClassifier} class is the method which provides the model text representation.

\code{.ctbn} format starts with the following lines:
\begin{CodeChunk}
\begin{Code}
-----------------------
BAYESIAN NETWORK
-----------------------
BBNodes N
-----------------------
\end{Code}
\end{CodeChunk}
where \code{N} is the number of nodes in the CTBNC (class node included).

Then each node is specified with the couple \code{<node\_name states\_number>} as follows:
\begin{CodeChunk}
\begin{Code}
Class 	4
N01	2
N02	2
N03	2
...
N15	4
-----------------------
\end{Code}
\end{CodeChunk}
where data in the same line are separated by spaces or tabs, and the first node is the class.

The next lines in the \code{.ctbn} format define the Bayesian Network (BN) structure, which represents the initial probability distribution of the CTBNC.
Each line starts with the considered node and then lists all its parents, using the space-separator format. Each line has to stop with a zero.
Here is an example:
\begin{CodeChunk}
\begin{Code}
Class	0
N01	Class	0
N02	Class	N01	0
N03	Class	0
...
N15	Class	N07	0
-----------------------
\end{Code}
\end{CodeChunk}
The \code{.ctbn} format allows to define the initial probability distribution by using a Bayesian Network.
On the contrary, the CTBNCToolkit does not support an initial probability distribution yet, but assumes an initial uniform distribution between the variables states. The only exception is for the class which has its own probability distribution, as specified later.
For this reason, for the moment the initial distribution is  represented as a disconnected Bayesian network.
\begin{CodeChunk}
\begin{Code}
Class 0
N01 0
N02 0
N03 0
...
N15 0
-----------------------
\end{Code}
\end{CodeChunk}
Note that lines composed of minus signs separate the different file sections.

The next lines inform about the conditional probability distributions of the defined BN. Each CPT is written following the ordering of the parent nodes defined in the previous section. The CPTs are separated by a line composed by minus signs.
Here is an example in case of uniform distribution for a disconnected BN:
\begin{CodeChunk}
\begin{Code}
Class
0 0 0 0 
-----------------------
N01
0.5 0.5 
-----------------------
...
-----------------------
N15
0.25 0.25 0.25 0.25 
-----------------------
\end{Code}
\end{CodeChunk}
where the line of the class probability distribution contains zeros, due to the fact that the class prior will be defined in the next section of the file format.
Since CTBNCToolkit currently supports only an initial uniform distribution, the CPTs shown are straightforward.
Nevertheless, the \code{.ctbn} format supports complex CPTs, where each line is the probability distribution setting the value of the parent set.
Considering the definition of the parent set of node \code{N02}:
\begin{CodeChunk}
\begin{Code}
N02	Class	N01	0
\end{Code}
\end{CodeChunk}
imaging that \code{Class} node is binary, node \code{N01} is ternary, and \code{N02} has 4 states. Node \code{N02} CPTs can be defined by the probability distribution lines in the following order:\\

\scalebox{1}{
	\addtolength{\tabcolsep}{-0pt}
	\begin{tabular}{|c|c|l|}
	\hline
	Class & N01 & probability distribution line \\
	\hline
	0 & 0	 & \code{0.2 0.3 0.1 0.4 }\\
	\hline
	1 & 0	 & \code{0.15 0.2 0.35 0.3 }\\
	\hline
	0 & 1	 & \code{0.42 0.08 0.23 0.27 }\\
	\hline
	1 & 1	 & \code{0.15 0.2 0.35 0.3 }\\
	\hline
	0 & 2	 & \code{0.2 0.3 0.1 0.4 }\\
	\hline
	1 & 2 	& \code{0.23 0.5 0.15 0.12 }\\
	\hline
\end{tabular}}\\

With the previous lines we defined the model variables and then the initial probability distribution.
In the next part of the file format the temporal model is defined.

First the graph structure is defined in the same way as previously done for the BN:
\begin{CodeChunk}
\begin{Code}
-----------------------
DIRECTED GRAPH 
-----------------------
Class	0
N01	Class	N05	0
N02	Class	N01	0
N03	Class	N01	0
...
N15	Class	N11	0
-----------------------
\end{Code}
\end{CodeChunk}
For each node its parents are specified.

Then the CIMs are defined for each node. Each line represents a complete CIM given the parent set instantiation.
The parent sets are defined iterating the parent values starting from the left as previously shown for the Bayesian Network CPTs.
\begin{CodeChunk}
\begin{Code}
-----------------------
CIMS
-----------------------
Class
0.2323008849557522 0.2488938053097345 0.2610619469026549 0.2577433628318584 
-----------------------
N01
-1.7437542178131549 1.7437542178131549 1.6677084125959616 -1.6677084125959616 
-1.1359184948768346 1.1359184948768346 1.468624833995604 -1.468624833995604 
-1.655441390834388 1.655441390834388 1.9083015334745217 -1.9083015334745217 
-1.992223211031885 1.992223211031885 1.128540290987483 -1.128540290987483 
-1.4597541436405608 1.4597541436405608 1.3875353532121044 -1.3875353532121044 
-1.8514888977211081 1.8514888977211081 1.2036550623637277 -1.2036550623637277 
-1.5536818371118852 1.5536818371118852 1.06784175008451 -1.06784175008451 
-1.2675007732387165 1.2675007732387165 1.8220901564788115 -1.8220901564788115 
-----------------------
\end{Code}
\end{CodeChunk}
It is worthwhile to note that for the class node the prior probability distribution is specified and not a temporal model since the class is a static node.

\subsection{Learning algorithms}\label{sec:learning-code}
\begin{figure}[h!]	
	\centering
		\includegraphics[width=1.05\textwidth]{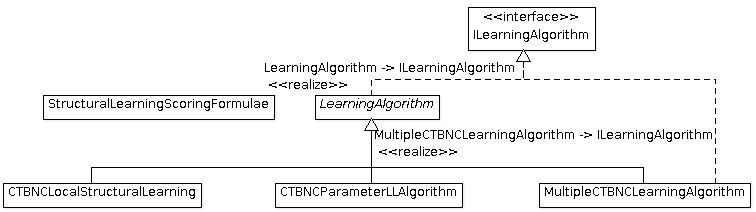}
	\caption{Learning algorithms simplified class diagram.}
	\label{fig:learningAlgorithms-class}
\end{figure}

\begin{figure}[h!]	
	\centering
		\includegraphics[width=1.0 \textwidth]{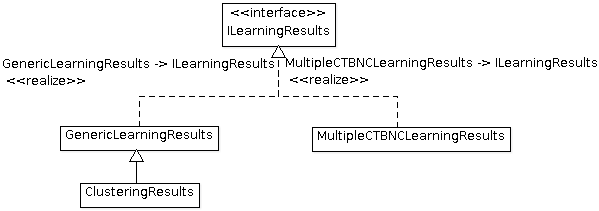}
	\caption{Simplified class diagram of the learning result components.}
	\label{fig:learningResults-class}
\end{figure}

Figures \ref{fig:learningAlgorithms-class} and \ref{fig:learningResults-class} depict the simplified class diagram of the components used to provide learning algorithms.

Learning algorithms are modeled by the \code{ILearningAlgorithm}interface, which defines the learning methods and the methods to manage the parameters of a learning algorithm.

\code{LearningAlgorithm} is an abstract class which provides an initial implementation of the parameter managing methods.
While \code{CTBNCParameterLLAlgorithm}, \code{CTBNCLocalStructuralLearning}, and \code{MultipleCTBNCLearningAlgorithm} are the learning algorithm implementations.

\code{CTBNCParameterLLAlgorithm} implements the parameter learning \cite{stella2012continuous}. This algorithm need in advance the definition of the CTBNC structure that specifies the dependencies between the variables. The parameters learning algorithm is the basis for all the structural learning algorithms.

\begin{figure}[h!]	
	\centering
		\includegraphics[width=1.3 \textwidth, angle=90]{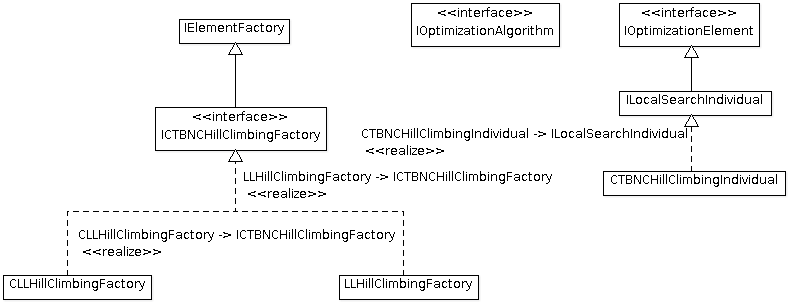}
	\caption{Hill climbing simplified class diagram.}
	\label{fig:hillclimbing-class}
\end{figure}

 \code{CTBNCLocalStructuralLearning} implements the structural learning for CTBNCs using the local search \citep{nodelman2002learning,codecasa13PKDD}. The searching algorithm relies on the optimization classes shown in Figure \ref{fig:hillclimbing-class}.

\code{MultipleCTBNCLearningAlgorithm} is the implementation of the local searching algorithm for \code{MultipleCTBNC} models (see Section \ref{sec:1vs1}). Its implementation is strictly related to the classical structural learning.

All the learning algorithms have to return the results of the learning process implemented by the \code{ILearningResults} interface.\\
\code{ILearningResults}  defines the learning results as a container for the sufficient statistics (i.e. \code{SufficientStatistics} class).

\code{GenericLearningResults} and \code{MultipleCTBNCLearningResults} provide a standard implementation of learning results, respectively for CTBNCs  and for \code{MultipleCTBNC} models.
\code{ClusteringResults} extends the \code{GenericLearningResults} class to implement the learning results for the clustering algorithm (see Section \ref{sec:clustering-code}).

Figure \ref{fig:hillclimbing-class} shows the simplified class diagram of the classes used for the local search in structural learning.

CTBNCs structural learning can be seen as an optimization problem. The CTBNCToolkit provides two scoring functions defined as \code{static} methods in \code{StructuralLearningScoringFormulae} class. The scoring functions rely on marginal log-likelihood  \citep{nodelman2002learning,codecasa13PKDD} and conditional log-likelihood calculation \citep{codecasa13PKDD}.

Since the learning procedure solves a maximization problem respect to a selected scoring function, the \code{IOptimizationElement} interface is defined. The interface provides a simple definition of an element that must be evaluated for optimization purposes. 
The \code{ILocalSearchIndividual} interface extends the \code{IOptimizationElement} interface, defining the properties of a local search element such as the ability to find the best neighbor (i.e. \code{getBestNeighbor} method).

\code{CTBNCHillClimbingIndividual} is the actual implementation of an individual in the local search procedure. It is the core of the local search algorithm implementation.

To generalize the local search algorithm, implemented in the \code{CTBNCLocalStructuralLearning} class, the factory pattern is used to generate local search individuals.

\code{IElementFactory} is the interface that defines the properties of a factory of \code{IOptimizationElement}. The use of Java generics helps the code generalization.
The \code{ICTBNCHillClimbingFactory}interface extends the \code{IElementFactory} interface to define the requirements for generating \code{CTBNCHillClimbingIndividual}.

\code{LLHillClimbingFactory} and \code{CLLHillClimbingFactory} are the two factories that generate \code{CTBNCHillClimbingIndividual}. The first factory generates the individuals for the marginal log-likelihood score maximization, while the second factory generates the individuals for the conditional log-likelihood score maximization.

\begin{example}
Here is an example of parameter learning.
\begin{CodeChunk}
\begin{Code}
// Model definition
CTBNClassifier clModel = new CTBNClassifier(nodeIndexing, "classifier",
	nodes);

// Structure definition
boolean[][] adjMatrix = new boolean[4][4];
for(int i = 0; i < adjMatrix.length; ++i)
	for(int j = 0; j < adjMatrix[i].length; ++j)
		adjMatrix[i][j] = false;
adjMatrix[0][1] = true;adjMatrix[0][2] = true;
adjMatrix[0][3] = true;
		
// Learning algorithm instantiation
CTBNCParameterLLAlgorithm  alg = new CTBNCParameterLLAlgorithm();

// Learning algorithm parameter (i.e. priors) definition
Map<String,Object> params = new TreeMap<String,Object>();
params.put("Mxx_prior", 1.0);
params.put("Tx_prior", 0.01);
params.put("Px_prior", 1.0);

// Learning algorithm parameters and structure setting
alg.setParameters(params);
alg.setStructure(adjMatrix);

// Parameter learning
Collection<ITrajectory<Double>> dataset = generateDataset();
alg.learn(clModel, dataset);
\end{Code}
\end{CodeChunk}
\end{example}

\begin{example}
Here is an example of structural learning using the marginal log-likelihood scoring.
\begin{CodeChunk}
\begin{Code}
// Parameter learning algorithm to use in the structural learning
CTBNCParameterLLAlgorithm  paramsAlg = new CTBNCParameterLLAlgorithm();
Map<String,Object> params = new TreeMap<String,Object>();
params.put("Mxx_prior", 1.0);
params.put("Tx_prior", 0.01);
params.put("Px_prior", 1.0);
paramsAlg.setParameters(params);

// Hill climbing factory definition
LLHillClimbingFactory elemFactory = new LLHillClimbingFactory(paramsAlg, 3,
	false, false);

// Definition of the local search starting structure
int iClass = nodeIndexing.getClassIndex();
int iA = nodeIndexing.getIndex("A");
int iB = nodeIndexing.getIndex("B");
int iC = nodeIndexing.getIndex("C");
boolean[][] adjMatrix = new boolean[4][4];
for(int i = 0; i < adjMatrix.length; ++i)
	for(int j = 0; j < adjMatrix[i].length; ++j)
		adjMatrix[i][j] = false;
adjMatrix[iClass][iA] = true;adjMatrix[iClass][iB] = true;
adjMatrix[iClass][iC] = true;

// Definition of the structural learning algorithm and setting of the initial
// structure
CTBNCLocalStructuralLearning<String,CTBNCHillClimbingIndividual> alg = 
	new CTBNCLocalStructuralLearning<String,CTBNCHillClimbingIndividual>(
		elemFactory);
alg.setStructure(adjMatrix);
		
// Structural learning of a target model
ICTClassifier<Double, CTDiscreteNode> model = generateClassifierModel();		
alg.learn(model, trainingSet);
boolean[][] learnedStructure = model.getAdjMatrix();
\end{Code}
\end{CodeChunk}
\end{example}

\subsection{Clustering}\label{sec:clustering-code}

\begin{figure}[h!]	
	\centering
		\includegraphics[width=0.75 \textwidth]{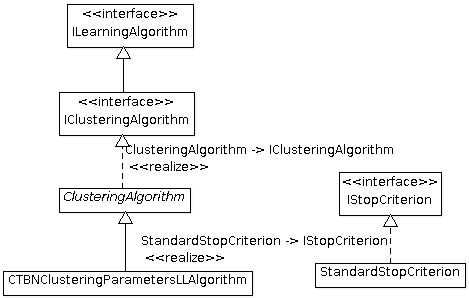}
	\caption{Simplified class diagram of the clustering learning.}
	\label{fig:clustering-class}
\end{figure}

Figure \ref{fig:clustering-class} depicts the simplified class diagram of the clustering learning.

\code{IClusteringAlgorithm} is the interface that defines the clustering learning algorithms. It extends the \code{ILearningAlgorithm} interface (Section \ref{sec:learning-code}).

\code{ClusteringAlgorithm} is an abstract class which implements the basis functionality of the clustering learning algorithm, while \code{CTBNClusteringParametersLLAlgorithm} implements the soft and hard assignment EM algorithms for clustering purposes \citep{codecasa14}. The structural learning algorithm relies on the optimization algorithms, used also in the case of supervised learning (Section \ref{sec:learning-code}).

\code{IStopCriterion} interface defines the stopping criterion for the EM iterative algorithm, and the \code{StandardStopCriterion} class provides a basic stop criterion implementation.

\begin{example}
Here is an example of soft-assignment EM clustering.
\begin{CodeChunk}
\begin{Code}
// Parameter definition
Map<String, Object> params = new TreeMap<String,Object>();
params.put("Mxx_prior", 1.0);
params.put("Tx_prior", 0.005);
params.put("Px_prior", 1.0);
params.put("hardClustering", false);
StandardStopCriterion stopCriterion = new StandardStopCriterion(
	iterationNumber, 0.1);

// Classification algorithm definition
Map<String, Object> paramsClassifyAlg = new TreeMap<String,Object>();
paramsClassifyAlg.put("probabilities", true);
CTBNCClassifyAlgorithm classificationAlg = new CTBNCClassifyAlgorithm();
classificationAlg.setParameters(paramsClassifyAlg);

// Clustering algorithm initialization
CTBNClusteringParametersLLAlgorithm cAlg =
	new CTBNClusteringParametersLLAlgorithm();
cAlg.setParameters(params);
cAlg.setClassificationAlgorithm(classificationAlg);
cAlg.setStopCriterion( stopCriterion);
		
// Clustering learning
ClusteringResults<Double> results = cAlg.learn(model, dataSet);
\end{Code}
\end{CodeChunk}
\end{example}

A classification algorithm is required for the expectation step of the EM algorithm in order to calculate the expected sufficient statistics (i.e. \code{CTBNCClassifyAlgorithm} in the previous code).

\subsection{Inference algorithms}\label{sec:inference-code}

\begin{figure}[h!]	
	\centering
		\includegraphics[width=0.7 \textwidth]{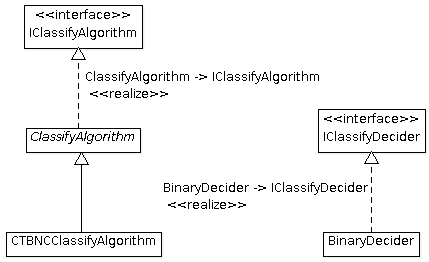}
	\caption{Simplified class diagram of inference components.}
	\label{fig:inferenceAlgorithms-class}
\end{figure}

Figure \ref{fig:inferenceAlgorithms-class} depicts the simplified class diagram of the inference algorithm classes.

\code{IClassifyAlgorithm} is the interface that defines the properties of an inference algorithm. It mainly defines properties related to parameter managing and classification methods. Classification methods return \code{IClassificationResult}.
As for the learning algorithms (Section \ref{sec:learning-code}), an abstract class is developed to manage the parameters, i.e. \code{ClassifyAlgorithm}. Generics are used to generalize the time representation and the nodes in the model to classify.

\code{CTBNCClassifyAlgorithm} implements the classification algorithm for CTBNCs, as described by \citet{stella2012continuous}.

\begin{example}
Here is an example of classification inference.
\begin{CodeChunk}
\begin{Code}
// Generate the trajectory to classify
ITrajectory<Double> testTrajectory = new CTTrajectory<Double>(nodeIndexing,
	times, values);

// Classification
CTBNCClassifyAlgorithm clAlgorithm = new CTBNCClassifyAlgorithm();
IClassificationResult<Double> result = clAlgorithm.classify(learnedModel,
	testTrajectory, 0.9);
\end{Code}
\end{CodeChunk}
\end{example}

Usually the classification relies on the most probable class, but defining a \code{IClassifyDecider} it is possible to use another classification criteria.
\code{BinaryDecider} implements the \code{IClassifyDecider} interface to allow an unbalanced classification in the case of two class problems. The decider allows to define a threshold to use in the classification in order to give an advantage to one of the two classes (see Section \ref{sec:bThreshold}).

\subsection{Validation methods}\label{sec:validation-code}

\begin{figure}[h!]	
	\centering
		\includegraphics[width=0.75 \textwidth]{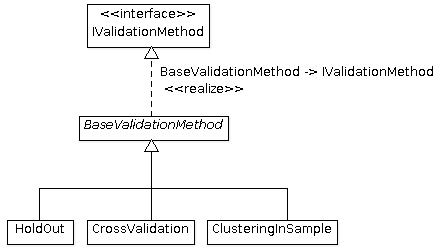}
	\caption{Simplified class diagram of the validation methods.}
	\label{fig:validationMethods-class}
\end{figure}

Figure \ref{fig:validationMethods-class} depicts the simplified class diagram of the validation methods used to realize tests and to calculate performances.

To execute the tests over a data set, validation methods are implemented. \code{IValidationMethod} is the interface that defines the validation method properties. \code{BaseValidationMethod} is the abstract class that implements the base functions of the validation methods.

\code{HoldOut}, \code{CrossValidation} and \code{ClusteringInSample} are the classes that implement different validation methods (Section \ref{sec:validation}). Each validation approach, one executed using \code{validate} method, returns the performances of the tested algorithm (Section \ref{sec:performances-code}).

\code{HoldOut} class implements the hold out validation, where training set and test set are used to calculate the performances.
\code{CrossValidation} class implements the cross validation method \citep{witten2005data}.
\code{ClusteringInSample} class implements a validation method, where learning and testing are both realized over the same complete data set; this can be used to test clustering approaches (Section \ref{sec:clustering}).

\begin{example}
Here is an example of classification inference.
\begin{CodeChunk}
\begin{Code}
// Instantiation of the cross validation method
CrossValidation<Double,CTDiscreteNode,CTBNClassifier,
	MicroMacroClassificationPerformances<Double,
		ClassificationStandardPerformances<Double>>,
	ClassificationStandardPerformances<Double>> validationMethod =
	new CrossValidation<...>(performanceFactory, 10, true);
validationMethod.setVerbose(true);

// Test execution using a validation method
performances = validationMethod.validate( model, learningAlgorithm,
	inferenceAlgorithm, dataset);
\end{Code}
\end{CodeChunk}
\code{performanceFactory} is a factory to generate new test performances. For more details about performances see Section \ref{sec:performances-code}.
\end{example}

\subsection{Performances}\label{sec:performances-code}
\begin{figure}[h!]	
	\centering
		\includegraphics[width=1.3\textwidth,angle=90]{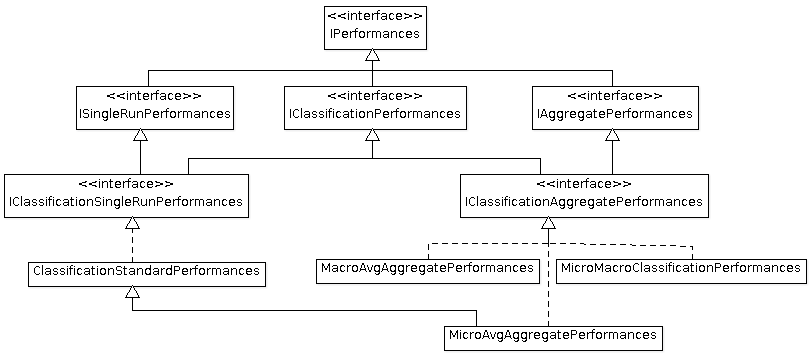}
	\caption{Simplified class diagram of the supervised classification performances.}
	\label{fig:performances-classification-class}
\end{figure}

\begin{figure}[h!]	
	\centering
		\includegraphics[width=1.3\textwidth,angle=90]{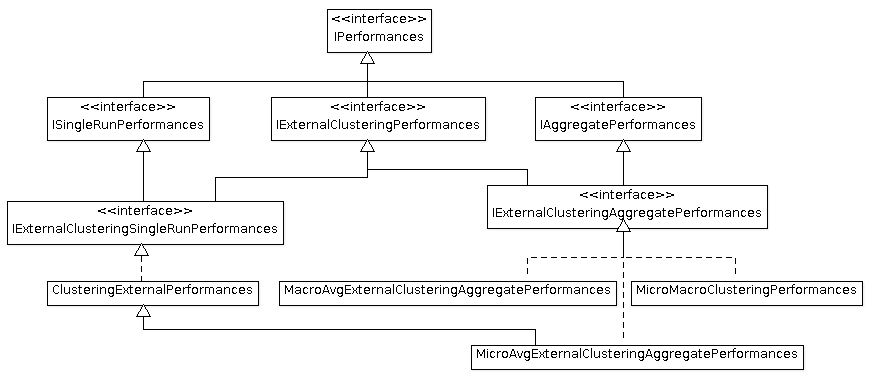}
	\caption{Simplified class diagram of the clustering performances.}
	\label{fig:performances-clustering-class}
\end{figure}

\begin{figure}[h!]	
	\centering
		\includegraphics[width=1\textwidth]{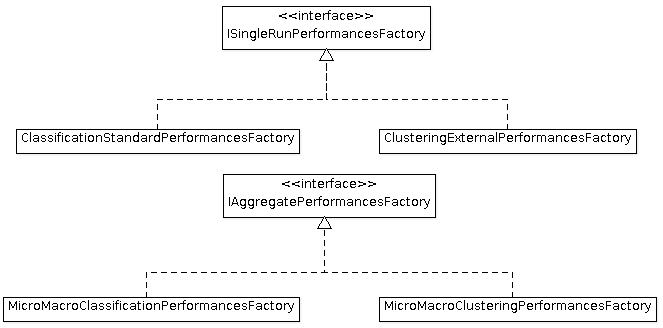}
	\caption{Simplified class diagram of the performance factories.}
	\label{fig:performances-factory-class}
\end{figure}

The CTBNCToolkit provides a rich set of performances to evaluate the experiments.  Different performances are provided in the case of classification (Section \ref{sec:results_class}) and clustering (Section \ref{sec:results_clust}).
The simplified class diagrams of the classification and the clustering performances are depicted in Figures \ref{fig:performances-classification-class} and \ref{fig:performances-clustering-class}.
In both cases the class hierarchy is the same. Different classes are provided to calculate single run and aggregate performances (i.e. for cross-validation multiple runs). To allow the best possible generalization, factory classes are provided to generate the performances. The simplified class diagram of the performance factories is depicted in Figure \ref{fig:performances-factory-class}. A factory argument is required in all the validation methods in order to generate the performances during the tests (see Section \ref{sec:validation-code}).

\code{IPerformances} is the interface which defines a generic performance, while \code{ISingleRunPerformances} and \code{IAggregatePerformances} respectively define the single run and the aggregate performances.

In the case of classification the \code{IClassificationPerformances} interface is provided. In the case of clustering the performances definitions relies on \code{IExternalClusteringPerformances}; only external clustering measures are provided (see Section \ref{sec:results_clust}).

\code{IClassificationSingleRunPerformances} is the interface which defines the single run classification performances, while the \code{IClassificationAggregatePerformances}interface defines the aggregate classification performances. 
Similarly, the clustering performances are defined by \code{IExternalClusteringSingleRunPerformances} and \code{IExternalClusteringAggregatePerformances} interfaces.

The \code{ClassificationStandardPerformances} class, together with the \code{ClusteringExternalPerformances} class, implements the performances for single runs in the case of classification and clustering. 
Aggregate performances are provided in micro and macro averaging. Micro averaging performances are calculated extending the single run performances. \code{MicroAvgAggregatePerformances} is the class that implements the micro averaging in the case of classification, while \code{MicroAvgExternalClusteringAggregatePerformances} implements the micro averaging performances in the case of clustering.

Macro averaging classification performances are provided by the \code{MacroAvgAggregatePerformances} class. Macro averaging performances for unsupervised learning (i.e. clustering) are provided by the \code{MarcoAvgExternalClusteringAggregatePerformances}  class.

In order to have the possibility to calculate both micro and macro averaging performances, two classes that hide both the averaging approaches are provided in the case of classification (i.e. \code{MicroMacroClassificationPerformances}) and in the case of clustering (i.e. \code{MicroMacroClusteringPerformances}).

Figure \ref{fig:performances-factory-class} depicts the simplified class diagram of the performance factory classes.

The \code{ISingleRunPerformancesFactory} interface defines the factory for single run performances. The factory for aggregate performances is defined by the \code{IAggregatePerformancesFactory} interface.
In the case of classification and in the case of clustering two factories are provided: one for the single run performances and one for the combination of micro and macro averaging aggregate performances. In the case of classification \code{ClassificationStandardPerformancesFactory} is provided for single runs, and \code{MicroMacroClassificationPerformancesFactory}is provided for aggregate performances. In the case of clustering the provided classes are: \code{ClusteringExternalPerformancesFactory} and \code{MicroMacroClusteringPerformancesFactory}.

\subsection{Tests}\label{sec:tests-code}
\begin{figure}[h!]	
	\centering
		\includegraphics[width=0.75 \textwidth]{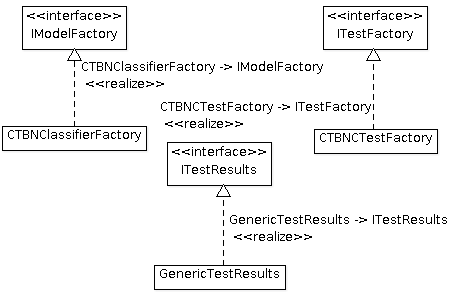}
	\caption{Simplified class diagram of the classes used to test CTBNCs.}
	\label{fig:tests-class}
\end{figure}

To perform the test experiments a set of utilities have been developed. Figure \ref{fig:tests-class} depicts the simplified class diagram of these utilities.

\code{IModelFactory} interface defines a factory to generate new models. \code{CTBNClassifierFactory} implements a simple way to generate synthetic CTBNCs. This class has been used to generate synthetic models for evaluation and test purposes.

\begin{example}
Here is an example of factory for the CTNB models.
\begin{CodeChunk}
\begin{Code}
// Number of variables
int N = 16;
// Number of states for each variable
int[] nStates = new int[N];
nStates[0] = 10; 		// class variable
nStates[1] = 2;  ...;  nStates[15] = 4;

// Range in which sample the lambda values of the exponential distribution
double[][] lambdaRanges = new double[2][N];
lambdaRanges[0][1] = 10; lambdaRanges[1][1] = 20;
...
lambdaRanges[0][15] = 40; lambdaRanges[1][15] = 80;
		
// Initialize the factory
CTBNClassifierFactory modelFactory = CTBNClassifierFactory( "naiveBayes",
	nStates, lambdaRanges);

// Generate a model
CTBNClassifier model = modelFactory.newInstance();
\end{Code}
\end{CodeChunk}
\end{example}

\code{ITestFactory} is an interface which defines a factory for tests. The idea is to provide a factory instance with the parameters of the tests, for example the learning and inference algorithms, and then to run a test calling a method. \code{CTBNCTestFactory} implements a test factory for CTBNCs. It provides the possibility to use a model factory to generate and test different data sets created from different model instances, and it provides the possibility to execute the tests over a data set in input.

In both cases \code{GenericTestResults} class is returned. This class implements a test result defined by the \code{ITestResults} interface and contains the performances of the tests.

\begin{example}
Here is an example of \code{CTBNCTestFactory} use. Generics are widely used to have a complete generalization.
\begin{CodeChunk}
\begin{Code}
// Factory instantiation
CTBNCTestFactory<Double,CTDiscreteNode,CTBNClassifier,
	MicroMacroClassificationPerformances<Double,
		ClassificationStandardPerformances<Double>>> testFactory;
testFactory = new CTBNCTestFactory<Double,CTDiscreteNode,CTBNClassifier,
	MicroMacroClassificationPerformances<Double,
		ClassificationStandardPerformances<Double>>>(
			modelFactory, validationMethod, nbParamsLearningAlg,
			classificationAlg);

// Test execution
GenericTestResults<Double,CTDiscreteNode,CTBNClassifier,
	MicroMacroClassificationPerformances<Double,
		ClassificationStandardPerformances<Double>>> resultNB = 
			testFactory.newTest("NB", datasetDim);

// Performances recovering
String perfSum = resultNB.performancesSummary(testName, confidenceLevel,
	datasetDim, kFolds);
\end{Code}
\end{CodeChunk}
\end{example}

In addition to the basic methods, \code{GenericTestResults} and  \code{CTBNCTestFactory} classes provide a set of general utilities to manage the tests.
\code{GenericTestResults} provides a set of \code{static} methods to print the classification and clustering performances.
\code{CTBNCTestFactory} provides a set of \code{static} methods to load and to manage the input data sets (see Section \ref{sec:input-code}).

\subsection{Command line front-end}\label{sec:frontend-code}
\code{CTBNCToolkit.frontend} is the package that contains the front-ends.
Currently, only a command line front-end is developed. The package consist of two classes: \code{Main} and \code{CommandLine}.
\code{Main} class is due to start the command line front-end implemented by the \code{CommandLine} class.

\subsubsection{Command line parameters}
The \code{CommandLine} class was developed to easily add new command line parameters (i.e. modifiers).

The command line parameters use an array of \code{String}s that gathers the information necessary to manage the modifiers (i.e. \code{modifiersList}).
\begin{CodeChunk}
\begin{Code}
private String[][] modifiersList = {
   {"help", "printHelp", "print the CTBNCToolkit help", "enabled"},
   ... ,
   {"validation", "setValidationMethod", "specify the validation method:\n" +
      "\t--validation=CV,k  \tk-folds cross validation is used\n" +
      "\t--validation=HO,0.6\thold out is used", "enabled"},
   ... ,
   {"v", "setVerbose", "enable the verbose comments", "enabled"}
};
\end{Code}
\end{CodeChunk}
Each line consists of four columns:
\begin{itemize}
	\item[i.] name of the modifier, i.e. \code{"help"} is the modifier name that can be called from the command line writing \code{-{}-help};
	\item[ii.] name of the function called to manage the modifier, i.e. when the \code{-{}-help} modifier is inserted, the \code{printHelp()} function is called;
	\item[iii.] modifier description, automatically printed when the \code{-{}-help} modifier is inserted;
	\item[iv.] flag to enable or not the modifier: if the fourth column contains \code{"enabled"} the modifier is enabled, otherwise it is completely ignored, even in the printing of the help.
\end{itemize}

When the command line is started, the input parameters are analyzed in order to call the appropriate methods.
It is possible to define a method without parameters, i.e. \code{printHelp()}:
\begin{CodeChunk}
\begin{Code}
public void printHelp() {
   System.out.println("CTBNCToolkit for classification");
   ....
}
\end{Code}
\end{CodeChunk}
or a method with parameters, i.e. \code{setValidationMethod( vMethod)}:
\begin{CodeChunk}
\begin{Code}
public void setValidationMethod( LinkedList<String> vMethod) {
   ....
}
\end{Code}
\end{CodeChunk}
The inputs are passed to the method through a linked list automatically generated from the command line parameters. The command line parameters with arguments have to be in the following form:
\begin{CodeChunk}
\begin{Code}
--modifier=arg1,arg2,..,argN
\end{Code}
\end{CodeChunk}
this generates a linked list of the following \code{String}s: \code{"arg1"}, \code{"arg2"}, ..., \code{"argN"}.
Using the example of the validation method (\ref{sec:validation}): the command line input \code{-{}-validation=HO,0.6} generates a linked list where \code{"HO"} is the first argument and \code{"0.6"} is the second.

Using the \code{modifiersList} and the automatic managing of the modifier parameters it is possible to add a new modifier just adding in the code a line in the \code{modifiersList} and the corresponding method, which can have in input any number of arguments.

\subsubsection{Program execution}
The command line parameters are used to set the parameters required during the program execution, while the real CTBNCToolkit starter relies on the \code{CommandLine()} constructor.

The constructor first initializes the modifier methods using \proglang{Java} reflexivity. This is done in the \code{initModifiers()} method.
It then analyzes the command line parameters, calling the right methods and checking the compatibilities between the parameters.

Once the modifiers are managed, the CTBNCToolkit starts with the following steps:
\begin{itemize}
	\item data set loading and generation of a CTNB, provided by the \code{loadDatasets(..)} method (see Section \ref{sec:input-code});
	\item loading of the learning algorithms to test, provided by the \code{loadLearningAlgorithms(..)} method (see Section \ref{sec:learning-code});
	\item loading of the classification algorithm, provided by the \code{loadInferenceAlgorithm()} method (see Section \ref{sec:inference-code});
	\item test generation, provided by the \code{generateTestFactories(..)} method (see Section \ref{sec:tests-code});
	\item test execution, provided by the \code{executeTests(..)} method (see Section \ref{sec:tests-code}).
\end{itemize}

%% file: conclusions.tex
\section{Conclusion and future works}\label{sec:conclusions}

In this paper the CTBNCToolkit has been presented. CTBNCToolkit is an open source toolkit for the temporal classification of a static variable using Continuous Time Bayesian Network Classifiers (CTBNCs).
Once introduced the main concepts about the CTBNCs \citep{stella2012continuous,codecasa13PKDD,codecasa14}, the stand-alone usage of the toolkit was provided.
The performances description and the tutorial examples allowed to replicate the examples and to test the toolkit, while the description of the \proglang{Java} library introduced the code giving the possibility to use the CTBNCToolkit as an external library for related application.

This paper describes the first version of CTBNCToolkit, for this reason many future developments are be planned. First the \code{-{}-model} (Section \ref{sec:model}) and the \code{-{}-testset} (Section \ref{sec:testset}) modifiers will be implemented.
The current version of CTBNCToolkit does not allow to the class node to have parents. One of the possible future implementation consists in the possibility to add static nodes as parents of another static node.

Further extension will be provided in parallel with the planned research on CTBNCs. Currently the research directions are focused on the possibility of extending the model in the case of partially observable trajectories.